\documentclass[preprint,12pt]{elsarticle}

\usepackage[utf8]{inputenc}
\usepackage[T1]{fontenc}
\usepackage{lmodern}
\usepackage{amsmath,amssymb,amsthm}
\usepackage{mathtools}
\usepackage{graphicx}
\usepackage{wrapfig}
\usepackage{booktabs}
\usepackage{enumitem}
\usepackage{microtype}
\usepackage[colorlinks=true,linkcolor=black,citecolor=blue,urlcolor=blue]{hyperref}
\usepackage{xcolor}
\usepackage{tikz}
\usepackage{pgfplots}
\pgfplotsset{compat=1.18}
\usetikzlibrary{positioning,arrows.meta,calc,backgrounds,fit,shapes.geometric}

\definecolor{qj10}{HTML}{D9ED92} 
\definecolor{qj20}{HTML}{B5E48C}
\definecolor{qj30}{HTML}{99D98C}
\definecolor{qj40}{HTML}{76C893}
\definecolor{qj50}{HTML}{52B69A} 
\definecolor{qj60}{HTML}{34A0A4} 
\definecolor{qj70}{HTML}{168AAD} 
\definecolor{qj80}{HTML}{1A759F}
\definecolor{qj90}{HTML}{1E6091}
\definecolor{qj100}{HTML}{184E77} 
\definecolor{qjink}{HTML}{0C1E2B} 
\definecolor{qjgrid}{HTML}{D5E1E6} 

\pgfplotsset{
  qjaxaxis/.style={
    width=7.4cm, height=5.6cm,
    axis line style={qjink!70, line width=0.5pt},
    tick label style={font=\scriptsize, color=qjink!75},
    label style={font=\small, color=qjink!85},
    title style={font=\small\bfseries, color=qjink!85},
    legend style={font=\scriptsize, draw=qjink!25, fill=white,
                  fill opacity=0.9, text opacity=1, rounded corners=1pt},
    grid=major, grid style={qjgrid, line width=0.3pt},
    tick align=outside, tick style={qjink!50},
    every axis plot/.append style={line width=1.1pt},
  },
}

\usepackage{listings}
\lstdefinestyle{qjpy}{
  language=Python,
  basicstyle=\ttfamily\scriptsize,
  keywordstyle=\color{qj100}\bfseries,
  commentstyle=\color{qjink!55}\itshape,
  stringstyle=\color{qj60},
  numberstyle=\tiny\color{qjink!45},
  emphstyle=\color{qj80},
  showstringspaces=false, columns=fullflexible, keepspaces=true,
  frame=leftline, framerule=0.8pt, rulecolor=\color{qj100!35},
  xleftmargin=8pt, aboveskip=5pt, belowskip=2pt, breaklines=true,
  emph={grad,value_and_grad,jit,sigmoid,softplus,log_softmax,gammaln},
}

\newcommand{\Sq}{S_q}
\newcommand{\lnq}{\ln_q}
\newcommand{\expq}{\exp_q}
\newcommand{\RR}{\mathbb{R}}

\DeclareMathOperator*{\argmax}{arg\,max}
\DeclareMathOperator*{\argmin}{arg\,min}
\DeclareMathOperator{\softmax}{softmax}
\DeclareMathOperator{\entmax}{entmax}

\DeclareMathOperator{\relu}{ReLU}

\newtheorem{theorem}{Theorem}

\journal{Physica A: Statistical Mechanics and its Applications}

\begin{document}

\begin{frontmatter}

\title{Perspectives on Tsallis Statistics for Artificial Intelligence}

\author[ucl,hai]{Kleyton da Costa}
\ead{kleyton.costa.25@ucl.ac.uk}

\author[utah]{Bernardo Modenesi}
\ead{bernardo.modenesi@utah.edu}

\affiliation[ucl]{organization={University College London},
                  city={London}, country={United Kingdom}}
\affiliation[hai]{organization={Holistic AI},
                  city={London}, country={United Kingdom}}
\affiliation[utah]{organization={Division of Biostatistics \& School of Computing, \\
                  University of Utah}, country={United States}}

\begin{abstract}
  
\noindent Tsallis statistics generalizes Boltzmann-Gibbs statistical mechanics through a single real parameter $q$ that controls the weight assigned to rare and frequent events. Originally proposed to describe physical systems with long-range correlations, multifractal geometry, and heavy-tailed fluctuations, the framework has become a recurring ingredient in modern artificial intelligence (AI): it underlies sparse attention mechanisms (\textsc{sparsemax} and $\alpha$-\textsc{entmax}), maximum-entropy reinforcement learning with controllable exploration, robust and heavy-tailed probabilistic models, and a family of generalized loss functions and regularizers. This paper offers a structured perspective on where Tsallis statistics meets AI. We first review the mathematical core: $q$-entropy and its variational (maximum-entropy) foundation, the $q$-exponential and $q$-logarithm, the $q$-central limit theorem, $q$-Gaussian distributions, and their dynamical origin in superstatistics, emphasizing the properties that matter for machine learning. We then survey applications across softmax generalization, reinforcement learning, sequential and graph neural models, generative and probabilistic modeling, loss design, and optimization, extracting the recurring design pattern in each case: a tunable interpolation between dense/uniform and sparse/peaked behavior governed by $q$. We further argue that the heavy-tailed weight spectra and gradient-noise statistics empirically observed in deep networks are themselves nonextensive signatures, placing modern learning dynamics within the scope of $q$-statistics. Finally, we discuss methodological pitfalls, the relationship to information geometry and $q$-exponential families, and open directions, arguing that $q$ should be treated as a learnable inductive bias rather than a fixed hyperparameter.
\end{abstract}


\end{frontmatter}

\section*{Highlights}
\begin{itemize}[nosep,leftmargin=1.2em]
  \item Reviews Tsallis nonextensive statistics as a unifying basis for AI.
  \item One entropic index $q$ deforms softmax, losses, priors and policies.
  \item Sparse attention, max-entropy RL, heavy-tailed models are
        $q$-deformations.
  \item Heavy-tailed weights and gradient noise are nonextensive signatures.
  \item Argues for treating $q$ as a learnable, context-dependent bias.
\end{itemize}

\section{Introduction}

\begin{figure}[t]
\centering
\begin{tikzpicture}[
    >={Latex[length=4.5pt,width=4.5pt]},
    font=\small,
    hub/.style={draw=black, line width=0.8pt, fill=black!5, align=center,
                inner xsep=6pt, inner ysep=5pt, rounded corners=1.5pt,
                font=\footnotesize},
    leaf/.style={draw=black!72, line width=0.5pt, fill=white, align=center,
                 inner sep=5pt, rounded corners=1.5pt, text width=3.05cm,
                 minimum height=1.45cm, font=\footnotesize},
    link/.style={draw=black!65, line width=0.7pt, -{Latex[length=4.5pt]},
                 shorten >=3pt, shorten <=3pt},
  ]
  \node[hub] (hub) at (0,0)
    {\textsc{Tsallis statistics}\\[3pt]
     $\Sq \;=\; \dfrac{1-\sum_i p_i^{\,q}}{q-1}$\\[3pt]
     \scriptsize $q\!\to\!1$:\ Boltzmann--Gibbs--Shannon};

  \node[leaf] (att) at (-5.0, 2.75)
    {\textsc{Sparse\\ attention}\\[3pt] $\alpha$-entmax$(\mathbf{z})$};
  \node[leaf] (gnn) at (-5.0, 0)
    {\textsc{Graph \&\\ sequence}\\[3pt] $\alpha$-entmax$_{\,j\in\mathcal{N}(i)}$};
  \node[leaf] (opt) at (-5.0,-2.75)
    {\textsc{Optimization}\\[3pt] $g_q(\Delta x)\!\propto\!\expq(-\beta\,\Delta x^2)$};
  \node[leaf] (rl) at (5.0, 2.75)
    {\textsc{Reinforcement\\ learning}\\[3pt] $\pi(a|s)\!\propto\!\expq\!\big(Q(s,a)/\tau\big)$};
  \node[leaf] (gen) at (5.0, 0)
    {\textsc{Generative\\ models}\\[3pt] $G_q(x)\!\propto\!\expq(-\beta x^2)$};
  \node[leaf] (loss) at (5.0,-2.75)
    {\textsc{Losses \&\\ regularizers}\\[3pt] $\mathcal{L}_q=-\sum_i y_i\,\lnq \hat p_i$};

  \foreach \n in {att,gnn,opt,rl,gen,loss}{\draw[link] (hub) -- (\n);}
\end{tikzpicture}
\caption{Tsallis statistics as an organizing field for artificial
intelligence. A single nonextensive core, the entropy $\Sq$ and its
$q$-deformed algebra, specializes through one entropic index $q$ into a
range of independently developed AI methods, each shown with the $q$-deformed
object it is built on: the Tsallis-entropy-regularized $\alpha$-entmax map for
sparse attention and sparse graph/sequence message passing; the $q$-exponential
policy of maximum-/sparse-entropy reinforcement learning; the $q$-Gaussian
density for heavy-tailed generative models; the $q$-cross-entropy (built from
the $q$-logarithm $\lnq$) for robust losses; and the heavy-tailed $q$-Gaussian
visiting distribution of generalized simulated annealing. Every arrow is a
``$q$-dial'': at $q=1$ the construction collapses to its
Boltzmann--Gibbs--Shannon counterpart (softmax, Boltzmann policy, Gaussian,
cross-entropy), while $q\neq1$ yields sparse or heavy-tailed behavior.}
\label{fig:field}
\end{figure}
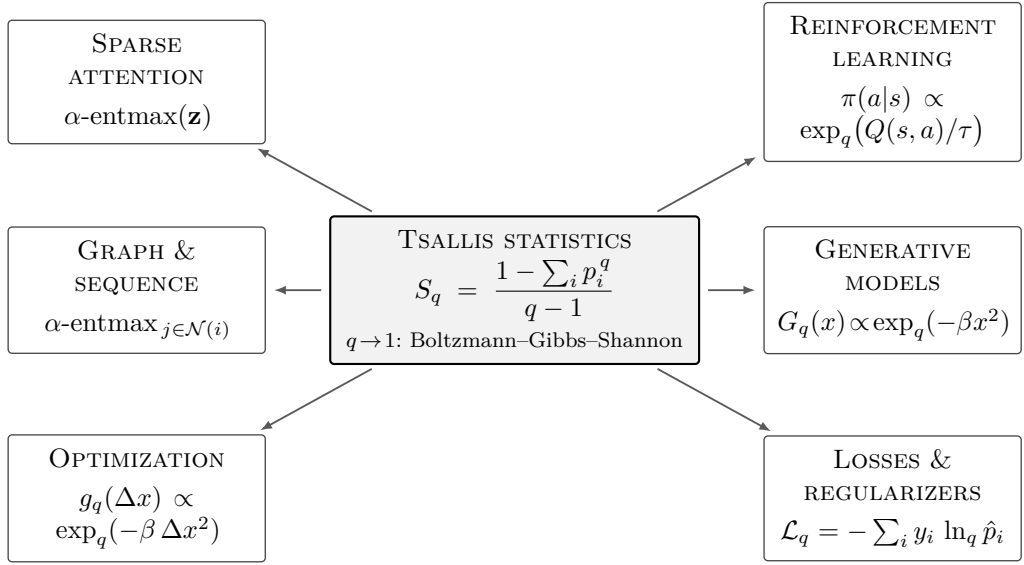

The dominant statistical backbone of modern deep learning models is the
Boltzmann-Gibbs-Shannon (BGS) framework. The softmax function, the
cross-entropy loss, the Gaussian likelihood, the KL divergence, and the
maximum-entropy principle are all consequences of assuming that information is
\emph{additive} and that uncertainty is measured by the Shannon entropy
$S = -\sum_i p_i \ln p_i$. This assumption is so deeply embedded that it is
rarely stated explicitly. It works extremely well when the underlying data are
weakly correlated, light-tailed, and effectively ergodic.

Many systems that AI is now asked to model violate these conditions. Natural
language exhibits Zipfian, heavy-tailed token statistics; financial,
geophysical, and network data show power-law fluctuations and long-range
dependence; attention distributions over long contexts are often genuinely
sparse rather than smoothly spread; and exploration in reinforcement learning
benefits from being tunable rather than fixed by the Shannon prescription. In
all of these cases the BGS machinery is not \emph{wrong} so much as
\emph{rigid}: it offers no dial with which to trade off the influence of rare
versus typical events.

Tsallis statistics provides exactly such a dial. Introduced
by~\citet{tsallis1988} as a possible generalization of Boltzmann--Gibbs
statistics, it replaces the Shannon entropy with the one-parameter family
\begin{equation}
  \Sq(p) \;=\; k\,\frac{1 - \sum_i p_i^{\,q}}{q - 1},
  \qquad q \in \RR,
  \label{eq:tsallis-entropy}
\end{equation}
which recovers the Shannon entropy in the limit $q \to 1$. The parameter $q$
is the \emph{entropic index}: values $q<1$ emphasize rare events (encouraging
spread, exploration, and heavy tails), while $q>1$ emphasize frequent events
(encouraging concentration, sparsity, and robustness to outliers). Around this
entropy grows a self-consistent ``$q$-algebra'' (the $q$-logarithm,
$q$-exponential, $q$-product, $q$-Gaussian, and a $q$-generalized central limit
theorem) that mirrors the BGS toolbox term by term and collapses onto it at
$q=1$.

\paragraph{Thesis of this paper.} Our central claim is that Tsallis statistics
is not merely a physical curiosity but a \emph{practical and unifying design
principle} for AI. A number of independently proposed methods, among them
sparse attention, sparse and maximum-entropy reinforcement learning,
heavy-tailed latent-variable models, robust regression losses, and focal-style
reweighting, can be read as instances of the same move: deform a BGS object
by a single parameter $q$ to obtain a tunable interpolation between
dense/uniform behavior ($q\to1$) and sparse/peaked or heavy-tailed behavior
($q\neq1$). Recognizing this shared structure clarifies the methods, exposes
their assumptions, and suggests that $q$ should be treated as a learnable
inductive bias.

\paragraph{Relation to existing reviews.} Entropy in machine learning has been
surveyed before, but with a different object in view.
\citet{sepulveda2024entropyreview} catalogue how a wide range of entropies
(Boltzmann--Gibbs, Shannon, von Neumann, Kolmogorov--Sinai, topological, and
others) are \emph{used as measures} in data analysis and learning, and
\citet{kumar2025entropyreview} review entropy measures across an even broader
set of fields. Both are organized by entropy functional and by application
domain. Our aim is narrower and, we think, complementary: we take a
\emph{single} deformation, follow it across AI, and argue that its parameter is
a design variable rather than a descriptive statistic. Where those reviews ask
which entropy best quantifies a given system, we ask what happens when one
entropy's index is placed inside a model and optimized. That difference in
posture is what motivates the emphasis here on maps and objectives
($\alpha$-entmax, $q$-losses, $q$-exponential policies) rather than on
estimators, and on the learnability of $q$ rather than on its measurement.

\paragraph{Scope and contributions.} This is a perspective and review paper.
We do not propose a new algorithm; instead we (i) give a self-contained,
ML-oriented review of the Tsallis toolkit
(Section~\ref{sec:overview}); (ii) survey and organize applications of Tsallis
statistics across the AI landscape, making the common ``$q$ as a dial'' pattern
explicit (Section~\ref{sec:applications}); (iii) discuss cross-cutting
methodological perspectives, including the link to information geometry and
$q$-exponential families, and common pitfalls
(Section~\ref{sec:perspectives}); (iv) outline open challenges and
research directions (Section~\ref{sec:future}); and (v) release
\texttt{qjax}, an open-source JAX library that implements the $q$-deformed
primitives reviewed here as differentiable functions in which $q$ is an
ordinary trainable argument, so that the paper's central claim can be executed
rather than only asserted (Appendix~\ref{app:qjax}).
Section~\ref{sec:conclusion} concludes.

\section{An Overview on Tsallis Statistics}
\label{sec:overview}
\setlength{\intextsep}{6pt}
\begin{wrapfigure}{r}{0.27\textwidth}
  \vspace{-\baselineskip}
  \includegraphics[width=\linewidth]{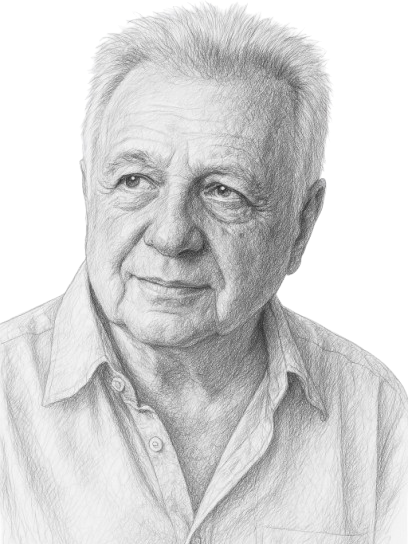}
  \caption{Constantino Tsallis, who introduced the generalized entropy
  $\Sq$ in 1988.}
  \label{fig:tsallis}
\end{wrapfigure}

This section reviews the mathematical core of Tsallis statistics at the level
of detail an ML researcher needs. Standard references are the founding
paper~\citep{tsallis1988}, the monograph~\citep{tsallis2009book}, and the
broader literature on generalized thermostatistics~\citep{naudts2011}. We work
throughout with discrete distributions $p=(p_1,\dots,p_n)$ on a finite alphabet
and with their continuous analogues where appropriate, and we set the Boltzmann
constant $k=1$ as is customary in information-theoretic usage.

\subsection{Tsallis entropy and nonextensive statistical mechanics}

\paragraph{The historical background.} In 1988, inspired by the geometry of
multifractals and by physical systems in which the Boltzmann--Gibbs formalism
fails (self-gravitating systems and stellar
polytropes~\citep{plastino1993}, long-range-interacting gases, turbulent
flows, and systems with long-lived memory), Constantino Tsallis proposed
the generalized entropy of Eq.~\eqref{eq:tsallis-entropy}. The proposal grew
into the field of \emph{nonextensive statistical mechanics}, with thousands of
applications across physics, biology, economics, and signal
processing~\citep{tsallis2009book,gellmann2004,tsallis2019beyondbgs}. The
defining departure from
BGS theory is the loss of additivity. For two probabilistically independent
subsystems $A$ and $B$ with joint distribution $p^{A\cup B}_{ij}=p^A_i p^B_j$,
the Tsallis entropy obeys the \emph{pseudo-additivity} rule
\begin{equation}
  \Sq(A \cup B) \;=\; \Sq(A) + \Sq(B) + (1-q)\,\Sq(A)\,\Sq(B),
  \label{eq:pseudo-additivity}
\end{equation}
so that the total entropy of the composite system is \emph{not} the sum of its
parts unless $q=1$. The sign of $(1-q)$ distinguishes \emph{superadditive}
($q<1$) from \emph{subadditive} ($q>1$) behavior, providing a coarse model of
correlation between subsystems that the additive Shannon entropy cannot
express.

\begin{figure}[t]
\centering
\includegraphics[width=\textwidth]{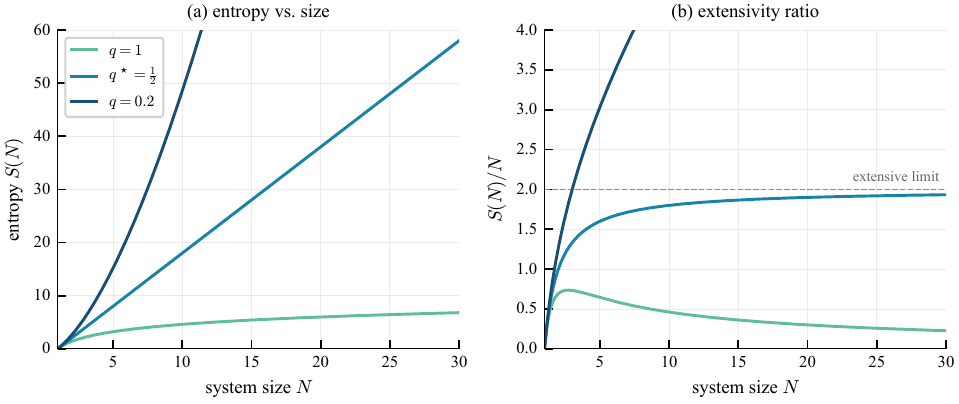}
\caption{Additivity is a property of the entropy functional; \emph{extensivity}
is a property of a particular system. For a strongly correlated system whose
accessible phase space grows as a power law $W(N)\propto N^{\rho}$ (here
$\rho=2$), the relevant quantity is the entropy's scaling with size $N$.
\textbf{(a)} The \emph{additive} Boltzmann--Gibbs entropy $S_{\mathrm{BG}}=\ln W$
grows only logarithmically, whereas the \emph{nonadditive} Tsallis entropy
$\Sq=\lnq W$ with the matching index $q^\ast=1-1/\rho=\tfrac12$ grows
\emph{linearly}; too small a $q$ over-corrects and grows faster than $N$.
\textbf{(b)} The extensivity ratio $S(N)/N$ makes the definition in
Eq.~\eqref{eq:extensivity} visual: it vanishes for $S_{\mathrm{BG}}$
(sub-extensive), converges to a finite constant only for $q^\ast$ (the unique
\emph{extensive} index), and diverges for $q=0.2$ (super-extensive). Only a
nonadditive entropy with the right $q$ recovers the linear scaling
thermodynamics requires.}
\label{fig:extensivity}
\end{figure}

\paragraph{Additivity versus extensivity.} These two properties are routinely
conflated but are logically independent, and keeping them apart is essential to
understanding why the theory is called \emph{nonextensive}. \emph{Additivity}
is a property of the entropy \emph{functional} alone: following
\citet{penrose1970}, $S$ is additive if
$S(A \cup B) = S(A) + S(B)$ for any two \emph{probabilistically independent}
systems, $p^{A\cup B}_{ij}=p^A_i p^B_j$. By this definition the
Boltzmann--Gibbs and the R\'enyi entropies are additive for every $q$, whereas
$\Sq$ is \emph{nonadditive} for all $q\neq1$, the residual term $(1-q)\Sq(A)\Sq(B)$
in Eq.~\eqref{eq:pseudo-additivity} being exactly the obstruction.
\emph{Extensivity}, by contrast, is a thermodynamic property of a \emph{specific
system}: for $\Sigma = A_1 \cup A_2 \cup \dots \cup A_N$ built from $N$ (not
necessarily independent) elements, $S$ is extensive if it grows linearly with
system size,
\begin{equation}
  0 \;<\; \lim_{N\to\infty}\frac{S(N)}{N} \;<\; \infty,
  \qquad\text{i.e.}\qquad S(N) \propto N \;\;(N\to\infty).
  \label{eq:extensivity}
\end{equation}
The two notions coincide only when the elements are independent or weakly
correlated. The point of nonextensive statistical mechanics is precisely the
strongly correlated case: when long-range interactions or memory make the
number of effectively accessible states grow sub-exponentially in $N$, the
\emph{additive} $S_{\mathrm{BG}}$ is \emph{not} extensive
($S_{\mathrm{BG}}(N)/N \to 0$), whereas the \emph{nonadditive} $\Sq$ with an
appropriate $q\neq1$ \emph{is} extensive and restores the linear scaling that
thermodynamics requires~\citep{tsallis2009book,tsallis2019beyondbgs}.
Figure~\ref{fig:extensivity} illustrates this for the canonical case of a
system whose number of accessible microstates grows as a power law
$W(N)\propto N^{\rho}$ (here $\rho=2$): the additive $S_{\mathrm{BG}}=\ln W$
grows only logarithmically and is sub-extensive, while $\Sq$ with the matching
index $q^\ast = 1-1/\rho$ grows linearly and is extensive; an over-deformed
choice ($q<q^\ast$) becomes super-extensive. Thus
nonadditivity is the microscopic signature of $\Sq$, while extensivity is the
macroscopic property it is designed to recover, which resolves the apparent
paradox in the field's name.

\paragraph{A hierarchy of extensive entropies.} The power-law case of
Figure~\ref{fig:extensivity} is only one row of a broader correspondence,
summarized in Table~\ref{tab:extensive-entropies}. For a system of $N$
equiprobable accessible states, the entropy of a single member of the family
evaluated on $W(N)$ states is, respectively,
\begin{equation}
  S_{\mathrm{BG}} = \ln W,\quad
  \Sq = \lnq W = \frac{W^{1-q}-1}{1-q},\quad
  S_\delta = (\ln W)^{\delta},\quad
  S_\lambda^{c}\ \text{(see below)} ,
  \label{eq:entropy-family}
\end{equation}
where $S_\delta = \sum_i p_i\,(\ln 1/p_i)^{\delta}$ and $S_\lambda^{c}$ are
further single-parameter deformations of the Shannon
form~\citep{tsallis2009book,tsallis2019beyondbgs}; all four reduce to
$S_{\mathrm{BG}}$ at their respective ``classical'' value
($q,\delta\to1$). The key observation is that \emph{which} member is extensive
is dictated entirely by how fast the phase space grows with system size:
\begin{itemize}[nosep]
  \item \emph{exponential} growth $W\sim\mu^{N}$, the ergodic, weakly
        correlated regime, makes the \emph{additive} $S_{\mathrm{BG}}$
        extensive, recovering standard thermodynamics;
  \item \emph{power-law} growth $W\sim N^{\rho}$, the strongly correlated
        regime of Figure~\ref{fig:extensivity}, makes $\Sq$ extensive for
        $q = 1-1/\rho$;
  \item \emph{stretched-exponential} growth $W\sim \nu^{N^{\gamma}}$ makes
        $S_\delta$ extensive for $\delta = 1/\gamma$;
  \item \emph{logarithmic} growth $W\sim D\ln N$, an extremely constrained
        phase space, makes $S_\lambda^{c}$ extensive for $\lambda = 1/D$.
\end{itemize}
The lesson is that $q$ is not privileged: it is one knob in a hierarchy of
entropies, each tuned so that the thermodynamically required linear scaling
$S\propto N$ is restored for a given class of correlations. Boltzmann--Gibbs is
simply the member matched to the (exponentially large, effectively
independent) phase spaces of textbook statistical mechanics.

\begin{table}[t]
\centering
\scriptsize
\setlength{\tabcolsep}{2pt}
\renewcommand{\arraystretch}{1.3}
\newcommand{\ext}[1]{\textcolor{qj80}{\textbf{extensive}}\,{\tiny(#1)}}
\newcommand{\extp}{\textcolor{qj80}{\textbf{extensive}}}
\newcommand{\nonext}{\textcolor{qjink!45}{non-ext.}}
\caption{Matching a generalized entropy to the growth law of the phase space.
For a system of $N$ elements with $W(N)$ equiprobable accessible states,
exactly one entropy in the hierarchy of Eq.~\eqref{eq:entropy-family} is
\emph{extensive} ($\,0<\lim_{N\to\infty}S/N<\infty$); the others are either
sub- or super-extensive. The required parameter value is shown in parentheses.
Additivity is fixed by the functional form ($S_{\mathrm{BG}}$ additive, the
rest nonadditive); extensivity, by contrast, depends on the system. Adapted
from~\citet{tsallis2009book,tsallis2019beyondbgs}.}
\label{tab:extensive-entropies}
\begin{tabular}{@{}l cccc@{}}
\toprule
 & \multicolumn{4}{c}{\textbf{Which entropy is extensive?}}\\
\cmidrule(lr){2-5}
\textbf{Phase-space growth } $W(N)$
  & $S_{\mathrm{BG}}$ & $\Sq$ & $S_\delta$ & $S_\lambda^{c}$\\
\textbf{(equiprobable)}
  & {\scriptsize additive} & {\scriptsize $q\neq1$}
  & {\scriptsize $\delta\neq1$} & {\scriptsize $\lambda>0$}\\
\midrule
exponential,\ \ $W\!\sim\!\mu^{N}$
  & \extp & \nonext & \nonext & \nonext\\
power law,\ \ $W\!\sim\! N^{\rho}$
  & \nonext & \ext{$q=1-\tfrac1\rho$} & \nonext & \nonext\\
stretched exp.,\ \ $W\!\sim\!\nu^{N^{\gamma}}$
  & \nonext & \nonext & \ext{$\delta=\tfrac1\gamma$} & \nonext\\
logarithmic,\ \ $W\!\sim\! D\ln N$
  & \nonext & \nonext & \nonext & \ext{$\lambda=\tfrac1D$}\\
\bottomrule
\end{tabular}
\end{table}

\paragraph{Equivalent forms and basic properties.} Using the $q$-logarithm
introduced below, Eq.~\eqref{eq:tsallis-entropy} can be written compactly as
\begin{equation}
  \Sq(p) \;=\; -\sum_i p_i^{\,q}\,\lnq p_i
            \;=\; \sum_i p_i\,\lnq\!\frac{1}{p_i},
\end{equation}
which makes the analogy with the Shannon form $S=-\sum_i p_i\ln p_i$
transparent. The Tsallis entropy is:
\begin{itemize}[nosep]
  \item \emph{nonnegative} for $q>0$ and concave in $p$ for $q>0$
        (and convex for $q<0$), so maximization remains
        well-posed~\citep{furuichi2006};
  \item \emph{maximized by the uniform distribution}, with maximum value
        $\Sq^{\max} = \lnq n = (n^{1-q}-1)/(1-q)$;
  \item \emph{reduced to Shannon entropy} as $q\to1$, by L'H\^opital's rule on
        Eq.~\eqref{eq:tsallis-entropy};
  \item the basis of a generalized maximum-entropy principle: maximizing $\Sq$
        subject to a normalization constraint and an energy-type expectation
        constraint yields a $q$-exponential (rather than exponential)
        distribution~\citep{tsallis1998}.
\end{itemize}

\paragraph{Related generalized entropies.} Tsallis entropy is monotonically
related to the R\'enyi entropy
$S^{R}_q = \tfrac{1}{1-q}\ln\sum_i p_i^q$ via
$S^{R}_q = \tfrac{1}{1-q}\ln\!\big[1+(1-q)\Sq\big]$; the two share the same
maximizers but differ in concavity and composition behavior. Both are members of the two-parameter Sharma--Mittal family, for which
closed-form expressions on exponential families are
known~\citep{nielsen2012qexp}, and of the broader class of $(h,\phi)$- and
trace-form entropies~\citep{naudts2011,hanel2011}. For machine learning the practically important
point is that $\Sq$ is a smooth, concave, single-parameter deformation of
Shannon entropy whose gradient is elementary, which is precisely what makes it
attractive as a regularizer.

\paragraph{Variational foundation and escort constraints.} The distributions of
nonextensive statistical mechanics arise, as in the Boltzmann--Gibbs case, from
a maximum-entropy principle. Extremizing $\Sq$ subject to normalization
$\sum_i p_i = 1$ and an energy constraint yields, through the Lagrangian
$\mathcal{L} = \Sq - \alpha\big(\sum_i p_i - 1\big)
- \beta\big(\langle\varepsilon\rangle_q - U_q\big)$,
a $q$-exponential equilibrium distribution
\begin{equation}
  p_i \;=\; \frac{1}{Z_q}\,\expq\!\big(-\beta\,(\varepsilon_i - U_q)\big),
  \qquad
  Z_q = \sum_j \expq\!\big(-\beta\,(\varepsilon_j - U_q)\big),
  \label{eq:qcanonical}
\end{equation}
the nonextensive analogue of the Boltzmann
weight~\citep{tsallis1998,curado1991}. A subtlety with no Boltzmann--Gibbs
counterpart is how the constraint expectation is defined. The internally
consistent choice~\citep{tsallis1998} uses the \emph{escort distribution}
\begin{equation}
  P_i \;=\; \frac{p_i^{\,q}}{\sum_j p_j^{\,q}},
  \qquad\text{so that}\qquad
  \langle\varepsilon\rangle_q \;=\; \sum_i P_i\,\varepsilon_i
  \;=\; \frac{\sum_i p_i^{\,q}\,\varepsilon_i}{\sum_j p_j^{\,q}} ,
  \label{eq:escort}
\end{equation}
which reweights the original distribution toward its high- ($q>1$) or
low-probability ($q<1$) region. Escort distributions are not a technicality:
they reappear throughout the theory: in the definition of the $q$-variance
used for the $q$-Gaussian, in the $q$-Fisher information of the
information-geometric picture (Section~\ref{sec:perspectives}), and, implicitly,
in the $p_i^{\,q}$ weighting that distinguishes the Tsallis entropy from the
Shannon entropy. Different constraint conventions (linear,
Curado--Tsallis, or normalized/escort) are related by transformations of
$\beta$ and yield the same family of $q$-exponential solutions, but mixing them
is a common source of inconsistency.

\subsection{$q$-exponential and $q$-logarithm functions}

The algebraic heart of the theory is a deformed pair of inverse functions. For
$x>0$ the \emph{$q$-logarithm} is
\begin{equation}
  \lnq x \;=\; \frac{x^{\,1-q} - 1}{1-q}, \qquad (q\neq1),
  \qquad \lim_{q\to1}\lnq x = \ln x,
  \label{eq:qlog}
\end{equation}
and its inverse, the \emph{$q$-exponential}, is
\begin{equation}
  \expq(x) \;=\; \big[\,1 + (1-q)\,x\,\big]_{+}^{\frac{1}{1-q}},
  \qquad \lim_{q\to1}\expq(x) = e^{x},
  \label{eq:qexp}
\end{equation}
where $[\,y\,]_{+} = \max(y,0)$ enforces the so-called \emph{Tsallis cut-off}:
when $1+(1-q)x \le 0$ the function is defined to be zero. This cut-off is the
source of \emph{sparsity}. For $q>1$ the $q$-exponential has power-law tails
$\expq(x)\sim x^{1/(1-q)}$ as $x\to\infty$, decaying polynomially rather than
exponentially; for $q<1$ it has compact support on the left, vanishing
identically once the argument is sufficiently negative.

\paragraph{Deformed algebra.} To preserve familiar identities one introduces
the \emph{$q$-sum} and \emph{$q$-product},
\begin{align}
  x \oplus_q y &= x + y + (1-q)\,xy, \\
  x \otimes_q y &= \big[\,x^{1-q} + y^{1-q} - 1\,\big]_{+}^{\frac{1}{1-q}},
\end{align}
which satisfy $\lnq(x\otimes_q y)=\lnq x + \lnq y$ and
$\expq(x)\,\expq(y)=\expq(x\oplus_q y)$, exactly mirroring the ordinary
logarithm and exponential. Eq.~\eqref{eq:pseudo-additivity} is just the
statement that Tsallis entropies compose under $\oplus_q$. These deformed
operations are what make the rest of the framework internally consistent: the
$q$-exponential family, the $q$-Fourier transform, and the $q$-CLT all rely on
them.

\subsection{$q$-Central limit theorem}

The ordinary central limit theorem (CLT) explains the ubiquity of the Gaussian:
normalized sums of \emph{independent} finite-variance random variables converge
to a Gaussian. Tsallis statistics is built for systems with \emph{strong
correlations}, where independence fails, and it comes with a matching limit
theorem. \citet{umarov2008} introduced the notion of \emph{$q$-independence},
a specific form of global correlation defined through the $q$-Fourier transform
and the $q$-product, and proved a generalized CLT. The construction rests on
the \emph{$q$-Fourier transform}
$F_q[f](\xi) = \int f(x)\,\expq\!\big(i\,\xi\,x\,[f(x)]^{q-1}\big)\,dx$,
which maps $q$-Gaussians to $q'$-Gaussians and linearizes the $q$-product, so
that the role played by the ordinary characteristic function in the classical
proof is taken over by its $q$-deformation~\citep{umarov2008,borges2004}. The
indices appearing at successive stages of the argument are linked by the
duality map
\begin{equation}
  q \;\longmapsto\; \frac{1+q}{3-q},
  \label{eq:qdual}
\end{equation}
which organizes a hierarchy of $q$-Gaussian attractors and reduces to the
identity at $q=1$.

\begin{theorem}[$q$-CLT, informal~\citep{umarov2008}]
Let $X_1, X_2, \dots$ be identically distributed random variables that are
$q$-independent in the above sense and have finite $(2q-1)$-variance. Then a
suitably rescaled sum $D_N^{-1}\sum_{i=1}^N X_i$ converges, as $N\to\infty$,
to a $q'$-Gaussian, whose index $q'$ is obtained from the correlation index $q$
through Eq.~\eqref{eq:qdual}. For $q=1$ one has $q'=1$ and
$D_N \propto \sqrt{N}$, and the statement reduces to the classical central
limit theorem.
\end{theorem}

Two qualifications are worth stating explicitly, since both are frequently
blurred in applied work. First, the exact result is formulated in terms of a
\emph{triplet} of indices that separately govern the attractor, the type of
correlation, and the scaling rate $D_N$; these are successive members of the
sequence generated by Eq.~\eqref{eq:qdual}, so the index of the limiting
$q$-Gaussian is in general \emph{not} the index describing the correlation
among the summands. Second, the relevant moment condition is finiteness of the
$(2q-1)$-variance rather than of the ordinary variance, which for $q>1$ is
strictly weaker; when that generalized variance diverges the attractor is no
longer a $q$-Gaussian but a $(q,\alpha)$-stable law, the nonextensive
counterpart of the L\'evy distributions~\citep{prato1999}. We refer
to~\citet{umarov2008} and~\citet{tsallis2009book} for the precise
bookkeeping.

The practical reading for AI is conceptual but important: \emph{heavy-tailed,
$q$-Gaussian limit laws are the natural attractors for aggregates of strongly
correlated variables}, just as the Gaussian is the attractor for independent
ones. Whenever a model aggregates correlated signals (pooled embeddings,
summed attention contributions, accumulated gradients with long-range
dependence), a $q$-Gaussian rather than Gaussian prior or noise model may
be the principled choice. The $q$-CLT thus provides theoretical grounding for
the heavy-tailed phenomena empirically observed in deep learning, such as the
heavy-tailed spectra of weight matrices and the heavy-tailed gradient noise
reported in stochastic optimization.

\subsection{$q$-Gaussian distributions}

Maximizing the Tsallis entropy under a fixed (generalized) variance constraint
yields the \emph{$q$-Gaussian}, the nonextensive analogue of the normal
distribution:
\begin{equation}
  G_q(x) \;=\; \frac{1}{Z_q}\,
  \expq\!\big(-\beta\,x^2\big)
  \;=\; \frac{1}{Z_q}\,
  \big[\,1 - (1-q)\,\beta\,x^2\,\big]_{+}^{\frac{1}{1-q}},
  \label{eq:qgaussian}
\end{equation}
where $\beta>0$ sets the width and $Z_q$ normalizes the density. The
$q$-Gaussian is a one-parameter bridge across several classical distributions:
\begin{center}
\begin{tabular}{@{}lll@{}}
\toprule
Regime & Distribution & Tail behavior \\
\midrule
$q\to1$        & Gaussian (normal)      & light (exponential) \\
$q=2$          & Cauchy / Lorentzian    & heavy (power law) \\
$1<q<3$        & Student-$t$ family     & power-law, normalizable \\
$q<1$          & compact support        & finite support (bounded) \\
$q\ge3$        & not normalizable       & n/a \\
\bottomrule
\end{tabular}
\end{center}
For $1<q<3$ the $q$-Gaussian is exactly a (rescaled) Student-$t$ distribution
with $\nu = (3-q)/(q-1)$ degrees of freedom, which is why so many ``robust''
and ``heavy-tailed'' models in statistics and ML are, implicitly, $q$-Gaussian
models. The variance is finite only for $q<5/3$; beyond that the distribution
is heavy enough that second moments diverge, and one works with the
\emph{$q$-variance} $\sigma_q^2 = \int x^2 [G_q(x)]^q\,dx / \int [G_q(x)]^q\,dx$
instead. This distribution, with a single tunable tail parameter, underlies most of the
generative and probabilistic uses of Tsallis statistics discussed in
Section~\ref{sec:applications}.

\begin{figure}[t]
\centering
\includegraphics[width=\textwidth]{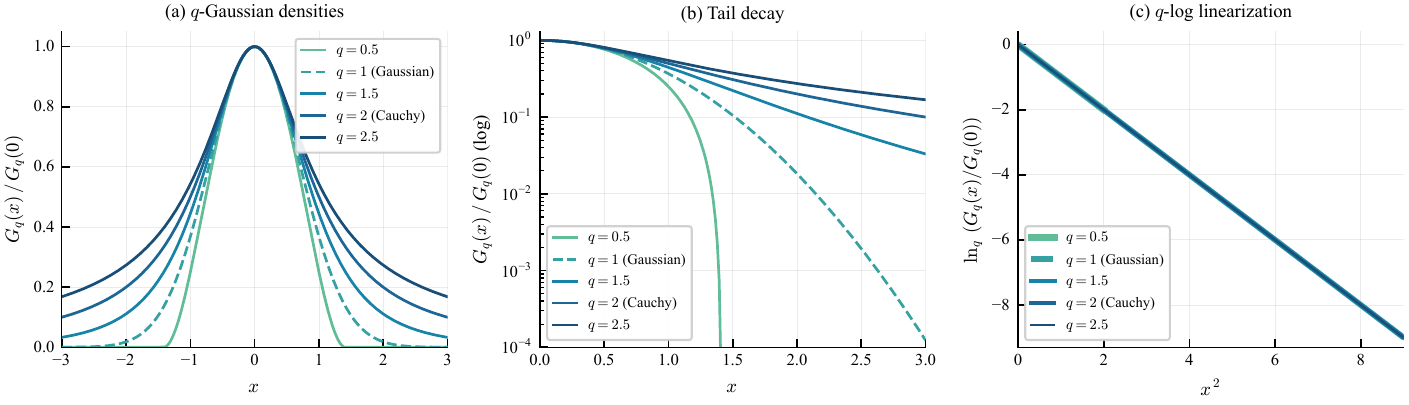}
\caption{The $q$-Gaussian $G_q(x)\propto\expq(-x^2)$ as a one-parameter bridge
across classical distributions (curves normalized to unit peak; all three panels
show the \emph{same} curves).
\textbf{(a)} On a linear scale, $q<1$ yields compact support (the curve
reaches exactly zero, via the Tsallis cut-off), $q=1$ is the Gaussian, and
$q>1$ produces progressively heavier tails ($q=2$ is the Cauchy
distribution). \textbf{(b)} On a semi-log scale the contrast is sharp: the
Gaussian ($q=1$) decays super-exponentially (a downward parabola), whereas
$q>1$ gives slowly-decaying power-law tails, the regime relevant to
heavy-tailed data and to the $q$-central limit theorem. \textbf{(c)} The
$q$-logarithm linearizes each curve: since $G_q(x)/G_q(0)=\expq(-x^2)$, plotting
$\lnq$ against $x^2$ (each curve with its own $q$) gives the straight line
$-x^2$, so all curves collapse onto a single line of slope $-1$ through the
origin, which is the operational signature of $q$-Gaussianity. The $q=0.5$ curve
terminates at $x^2=1/(1-q)=2$, where its compact support ends.}
\label{fig:qgauss}
\end{figure}

\subsection{Superstatistics and the dynamical origin of $q$}
\label{sec:superstat}

A question that recurs whenever $q$-statistics is applied is \emph{where the
index $q$ comes from} physically. The most transparent answer is given by
\emph{superstatistics}~\citep{beck2003}, the statistics of a system whose
intensive parameter (typically an inverse temperature $\beta$) is itself a
slowly fluctuating random variable. If, conditionally on $\beta$, the system is
locally Boltzmannian, $p(\varepsilon\mid\beta)\propto e^{-\beta\varepsilon}$,
then the marginal distribution is the mixture
\begin{equation}
  p(\varepsilon) \;=\; \int_0^\infty f(\beta)\,e^{-\beta\varepsilon}\,d\beta,
  \label{eq:superstat}
\end{equation}
and the particular choice of a $\chi^2$ (gamma) distribution for $f(\beta)$
reproduces \emph{exactly} the $q$-exponential of Eq.~\eqref{eq:qexp}, with the
index fixed by the relative variance of the temperature fluctuations,
$q = 1 + \mathrm{Var}(\beta)/\langle\beta\rangle^2$. Nonextensivity thus emerges
from a hidden hierarchy of scales: $q>1$ is a direct measure of the
heterogeneity of the local environments that have been averaged over.

This dynamical picture is the natural bridge to machine learning, where mixtures
over fluctuating ``temperatures'' are ubiquitous. Mini-batch stochastic gradient
descent samples a different local loss landscape at each step; ensembles, dropout
and Bayesian posteriors average over fluctuating parameters; and mixture-of-experts
and attention models superpose components with different effective scales. To the
extent that these fluctuations are gamma-like, Eq.~\eqref{eq:superstat} predicts
$q$-exponential and $q$-Gaussian statistics for the aggregated quantities,
a prediction we take up in Section~\ref{sec:dynamics}. Superstatistics also
supplies a constructive recipe for \emph{choosing} $q$ from the measured
variability of a scale parameter, rather than fitting it blindly.

\section{Applications of Tsallis Statistics in Artificial Intelligence}
\label{sec:applications}

We now survey how the toolkit of Section~\ref{sec:overview} appears across AI.
A recurring pattern, which we flag throughout, is what we call the
\emph{$q$-dial}: a method takes a standard BGS object (softmax, Shannon-entropy
regularizer, Gaussian prior, cross-entropy loss) and deforms it by $q$ to
obtain a controllable interpolation between dense and sparse, or light-tailed
and heavy-tailed, behavior. Table~\ref{tab:summary} summarizes the mapping.

\begin{table}[t]
\centering
\small
\renewcommand{\arraystretch}{1.25}
\begin{tabular}{@{}p{2.9cm}p{3.1cm}p{2.8cm}p{3.6cm}@{}}
\toprule
\textbf{Area} & \textbf{BGS baseline} & \textbf{Tsallis object} &
\textbf{Effect of the $q$-dial} \\
\midrule
Attention / output &
softmax, Shannon entropy &
$\alpha$-entmax, $\Sq$ regularizer &
$q\!=\!1$ dense softmax $\to$ $q\!>\!1$ sparse, exactly-zero weights \\
Reinforcement learning &
Shannon max-ent (SAC) &
Tsallis max-ent MDP &
tunes exploration; $q\!>\!1$ yields sparse, near-greedy policies \\
Generative / latent models &
Gaussian prior/likelihood &
$q$-Gaussian, Student-$t$ &
heavier tails, outlier robustness, mode coverage \\
Loss functions &
cross-entropy, MSE &
Tsallis / $q$-cross-entropy, $q$-loss &
robustness to noisy labels and outliers \\
Optimization &
gradient descent &
$q$-gradient, $q$-Gaussian search &
non-local steps, escaping local minima \\
\bottomrule
\end{tabular}
\caption{The ``$q$-dial'' pattern: representative areas where a
Boltzmann--Gibbs--Shannon object is replaced by its Tsallis deformation, with
the qualitative effect of moving $q$ away from $1$.}
\label{tab:summary}
\end{table}

\subsection{Softmax generalization (entmax and $\alpha$-softmax)}

The clearest and most influential AI application of Tsallis statistics is the
generalization of the softmax. The standard softmax,
$\softmax(z)_i = e^{z_i}/\sum_j e^{z_j}$, is the solution of the
entropy-regularized problem
$\argmax_{p\in\Delta}\, p^\top z + H(p)$, where $\Delta$ is the probability
simplex and $H$ is the Shannon entropy. Because Shannon entropy is steep at the
boundary of the simplex, the optimal $p$ always assigns \emph{strictly
positive} mass to every coordinate: softmax is irreducibly dense.

Replacing $H$ with the Tsallis entropy $\Sq$ yields the \emph{Tsallis entmax}
transformation~\citep{martins2016sparsemax,peters2019sparse}:
\begin{equation}
  \alpha\text{-}\entmax(z)
  \;=\; \argmax_{p\in\Delta}\; p^\top z + \Sq(p),
  \qquad \alpha = q,
  \label{eq:entmax}
\end{equation}
where the entmax literature writes the entropic index as $\alpha$.\footnote{The
two literatures normalize the regularizer differently.
\citet{peters2019sparse} use the Tsallis entropy in the form
$H^{T}_\alpha(p)=\frac{1}{\alpha(\alpha-1)}\sum_j\big(p_j-p_j^{\alpha}\big)$,
which on the simplex equals $\Sq(p)/\alpha$. With the normalization of
Eq.~\eqref{eq:tsallis-entropy} used throughout this paper, the right-hand side
of Eq.~\eqref{eq:entmax} therefore returns $\entmax_q(z/q)$ rather than
$\entmax_q(z)$, so the identifications below hold up to this rescaling of the
scores. We keep the $\Sq$ normalization for internal consistency and flag the
factor because mixing the two conventions is an easy source of error.} Two
special cases are central:
\begin{itemize}[nosep]
  \item $\alpha=1$ recovers the dense \textbf{softmax};
  \item $\alpha=2$ gives \textbf{sparsemax}~\citep{martins2016sparsemax}, a
        Euclidean projection onto the simplex that returns \emph{exactly zero}
        probabilities for low-scoring options;
  \item general $\alpha>1$ gives \textbf{$\alpha$-entmax}~\citep{peters2019sparse},
        with $\alpha=1.5$ a popular, differentiable middle ground.
\end{itemize}
The sparsity is a direct consequence of the Tsallis cut-off in
Eq.~\eqref{eq:qexp}: the KKT conditions of~\eqref{eq:entmax} threshold the
scores, so coordinates below a data-dependent threshold are clipped to zero.
This makes attention and output distributions \emph{interpretable} (the model
commits to a small support) and is exploited in sparse sequence-to-sequence
models, where exactly-zero output probabilities accelerate decoding and yield
crisp alignments~\citep{peters2019sparse}. \citet{correia2019adaptively}
extended this to \emph{adaptively sparse Transformers}, learning a separate
$\alpha$ per attention head, so that each head discovers how sparse it should
be, an early instance of treating $q$ as a learnable parameter. The entmax
loss (the Fenchel--Young loss induced by $\Sq$) provides a well-behaved
training objective with the entmax map as its
gradient~\citep{blondel2020fy}. The same Tsallis-entropy construction has since
been extended from the simplex to \emph{continuous} domains, giving sparse and
continuous attention densities and a corresponding family of Fenchel--Young
losses~\citep{martins2020sparse,martins2022sparse}.

Two more recent developments remove the main practical obstacles to using these
maps at scale. On the systems side, \citet{goncalves2025adasplash} give a
GPU-efficient $\alpha$-entmax implementation combining a hybrid
Halley--bisection solver with Triton kernels that exploit the induced sparsity,
which closes much of the runtime and memory gap to dense fused-attention
baselines on RoBERTa, ModernBERT, and GPT-2 scale models. On the modeling side,
\citet{vasylenko2026longcontext} make the entropic index and an attention
temperature jointly learnable and sequence-length dependent, and report that
the resulting adaptively sparse attention extrapolates well beyond its training
length in regimes where softmax attention disperses. Both bear directly on the
open questions of Section~\ref{sec:future}: the $q$-dial is now cheap enough to
run at scale, and the reported gains come from letting the data set $q$ rather
than from fixing it in advance.

\begin{figure}[t]
\centering
\includegraphics[width=\textwidth]{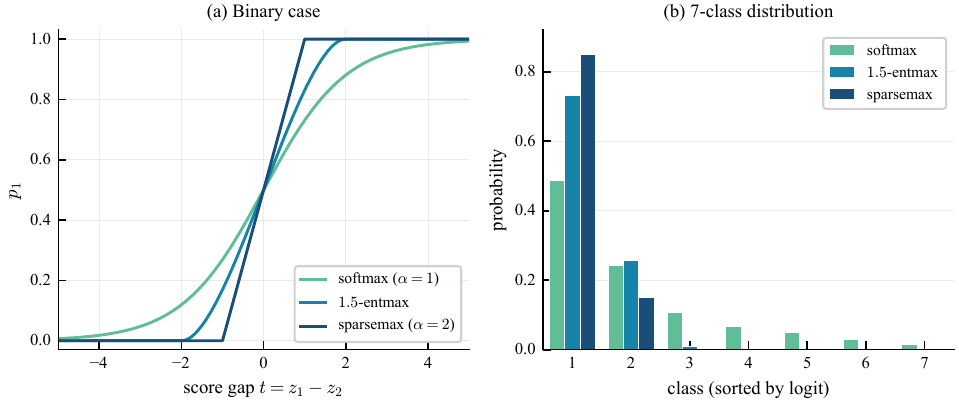}
\caption{Generalizing the softmax via Tsallis entropy
(Eq.~\eqref{eq:entmax}). \textbf{(a)} For two classes with score gap $t$, the
softmax ($\alpha=1$) is a smooth sigmoid that never reaches $0$ or $1$, so
every class keeps positive mass; $\alpha$-entmax with $\alpha>1$ saturates at a
finite margin, assigning \emph{exactly} $0$ or $1$ (sparsemax, $\alpha=2$,
saturates earliest, at $|t|=1$). \textbf{(b)} On a $7$-class example with logits
$z=(2.5,1.8,1.0,0.5,0.2,{-}0.3,{-}1.0)$, softmax spreads mass over all classes,
$1.5$-entmax keeps three, and sparsemax keeps only two, with the rest exactly
zero. The entropic index $\alpha=q$ is a continuous dial from dense to sparse.}
\label{fig:entmax}
\end{figure}

\subsection{Reinforcement learning}

Maximum-entropy reinforcement learning (RL) augments the reward with an entropy
bonus to encourage exploration and robustness; soft
actor--critic~\citep{haarnoja2018sac} and soft $Q$-learning are the canonical
Shannon-entropy instances. Replacing Shannon
entropy with Tsallis entropy gives \emph{Tsallis-entropy-regularized
MDPs}~\citep{lee2018sparse,lee2020tsallis,chow2018tsallis}, in which the
optimal policy is a $q$-exponential (rather than Boltzmann) function of the
$Q$-values:
\begin{equation}
  \pi^*(a\mid s) \;\propto\; \expq\!\big(Q(s,a)/\tau\big).
\end{equation}
The entropic index $q$ becomes a knob on the exploration--exploitation
trade-off and on policy sparsity:
\begin{itemize}[nosep]
  \item $q\to1$ recovers the standard \textbf{soft (Boltzmann) policy}, which
        is dense: every action retains positive probability;
  \item $q=2$ gives the \textbf{sparse MDP} of~\citet{lee2018sparse}, whose
        optimal policy assigns exactly zero probability to clearly suboptimal
        actions, much like sparsemax acts in the supervised setting;
  \item large $q$ approaches a \textbf{greedy} policy.
\end{itemize}
Tsallis regularization yields a generalized ``sparse Bellman'' operator with its
own contraction and convergence guarantees~\citep{lee2020tsallis}, and the path-
consistency learning framework extends to the Tsallis case~\citep{chow2018tsallis}.
More broadly, these methods are instances of \emph{regularized} MDPs, for which a
general Legendre--Fenchel theory clarifies how any convex (Shannon, Tsallis, or
KL) regularizer shapes the optimal policy and its error
propagation~\citep{geist2019regularizedmdp,vieillard2020munchausen}; the same
Tsallis-entropy idea also yields best-of-both-worlds optimal regret in the
bandit setting~\citep{zimmert2021tsallisinf}. The benefits reported include more
stable learning, automatic pruning of bad actions in large discrete action
spaces, and the ability to tune how committed a policy is by changing only $q$
rather than the algorithm.

\subsection{Sequential models}

Sequence models inherit the softmax generalization directly. In
sequence-to-sequence translation and language modeling, replacing the output
softmax and the attention softmax with $\alpha$-entmax produces sparse, peaked
attention and sparse next-token distributions, which improve interpretability
and can sharpen alignments while leaving accuracy unchanged or slightly
improved~\citep{peters2019sparse,correia2019adaptively}. Beyond
the softmax, the $q$-CLT and $q$-Gaussian perspective is relevant to sequence
models in two further ways. First, the heavy-tailed nature of token-frequency
distributions (Zipf's law) is naturally described in the $q$-exponential family,
suggesting $q$-deformed output parameterizations for rare-token modeling.
Second, the heavy-tailed gradient noise observed when training recurrent and
Transformer models on long sequences is consistent with $q$-Gaussian rather
than Gaussian statistics, which has implications for the choice of optimizer
and for the analysis of convergence. This is more than an analogy:
\citet{zhang2020adaptiveattention} show that attention models generate
heavy-tailed gradient noise \emph{independently of the input data}, and that
this, rather than curvature, is what makes adaptive and clipped methods
outperform plain SGD on such architectures. A heavy-tailed noise law that is
induced by the architecture itself is precisely the situation the $q$-CLT
describes, and it suggests that the appropriate noise model in convergence
analyses of Transformer training is $q$-Gaussian rather than Gaussian.
Sparsity in trained attention maps is also predictable in advance well enough
to be exploited computationally~\citep{treviso2022predicting}. Generalized
$q$-deformed state-space and attention kernels are an active, if still nascent,
direction.

\subsection{Graph neural networks}

Graph attention networks compute attention weights over a node's neighborhood
with a softmax, producing dense weights even when only a few neighbors are
relevant. Substituting $\alpha$-entmax for softmax in the neighborhood
aggregation yields \emph{sparse graph attention}: each node attends to a
learned, sparse subset of its neighbors, with the rest receiving exactly zero
weight. This exposes which edges the model actually uses and provides a
built-in, differentiable form of edge pruning. The motivation is well
documented even where the mechanism is not entmax:
\citet{ye2021sgat} show that attention sparsified by an $L_0$ penalty removes a
large fraction of task-irrelevant edges on standard benchmarks without loss of
accuracy, which is the same objective an entropic index attains through the
Tsallis cut-off rather than through a separate penalty term. The connection to
over-smoothing should be stated more carefully than it usually is. That
repeated aggregation drives node representations toward a low-dimensional
subspace is established~\citep{li2018deeperinsights,oono2020oversmoothing}, and
zeroing a subset of neighbors plainly changes the operator being iterated, but
we are not aware of a result showing that entmax aggregation provably slows
that convergence. We therefore record it as a plausible and testable
conjecture rather than a demonstrated benefit, and note that quantifying it
would be a concrete contribution. As in the Transformer case, learning a
per-layer or per-head $\alpha$ lets the network decide how aggressively to
sparsify its message passing. More broadly, the heavy-tailed degree
distributions characteristic of real-world networks are themselves
$q$-exponential, so nonextensive statistics offers a matched generative
description of the graphs that GNNs operate on.

\subsection{Generative and probabilistic models}

Wherever a Gaussian assumption appears in a probabilistic model, the
$q$-Gaussian of Eq.~\eqref{eq:qgaussian} offers a heavier-tailed,
outlier-robust alternative with a single extra parameter. Concretely:
\begin{itemize}[nosep]
  \item \textbf{Latent-variable models.} Replacing the Gaussian prior or
        posterior of a variational autoencoder with a $q$-Gaussian (Student-$t$)
        yields heavier-tailed latent spaces that better capture outliers and
        improve robustness; the Student-$t$ VAE~\citep{takahashi2018studentt}
        is, in this language, a $q$-Gaussian VAE.
  \item \textbf{Diffusion and score-based models.} The forward noising process
        need not be Gaussian; heavy-tailed ($q$-Gaussian / $\alpha$-stable)
        noise and other non-Gaussian, exponential-family corruptions have been
        explored to better model data with rare, large excursions and to
        improve robustness of the reverse
        process~\citep{okhotin2023starshaped,pandey2024heavytailed,shariatian2024dlpm}.
  \item \textbf{Robust density estimation and clustering.} Mixtures of
        $q$-Gaussians (Student-$t$ mixtures) downweight outliers automatically
        relative to Gaussian mixtures, a long-standing benefit in robust
        statistics that the Tsallis framework explains through the entropic
        index.
  \item \textbf{Embeddings and visualization.} $t$-SNE~\citep{vandermaaten2008tsne}
        uses a Student-$t$ ($q$-Gaussian with $q=2$) kernel in the
        low-dimensional space to obtain the heavy tails that relieve the
        crowding problem; the method is, at its core, a $q$-Gaussian
        construction.
\end{itemize}
The common thread is again the $q$-dial: $q\to1$ recovers the familiar
light-tailed Gaussian model, while $q>1$ trades heavier tails for robustness
and mode coverage.

\subsection{Loss functions and regularization}

Tsallis statistics motivates a family of generalized losses and regularizers,
all obtained by deforming a logarithm, an entropy, or a divergence with a single
index $q$. In every case $q$ trades off fidelity to confident predictions
against robustness to contamination, recovering the standard objective as
$q\to1$.

\paragraph{The $q$-cross-entropy.} The \emph{Tsallis (or $q$-)cross-entropy} is defined as
\begin{equation}
  \mathcal{L}_q = -\sum_i y_i\,\lnq \hat p_i ,
  \label{eq:q-cross-entropy}
\end{equation}
which interpolates between the standard cross-entropy ($q\to1$) and, for
$q\neq1$, losses that are more tolerant of noisy labels: the $q$-logarithm
bounds the penalty assigned to confidently wrong predictions. Following the
standard loss convention, $\mathcal{L}_q$ weights $\lnq \hat p_i$ by the label $y_i$
(not $y_i^{\,q}$).

\paragraph{Why the loss is not the entropy.} Unlike the Shannon case, the
diagonal of $\mathcal{L}_q$ does \emph{not} recover the Tsallis entropy, since
\begin{equation}
  -\sum_i p_i\,\lnq p_i \;\neq\; -\sum_i p_i^{\,q}\,\lnq p_i \;=\; \Sq(p)
  \qquad (q\neq1).
\end{equation}
The reason is that the $q$-logarithm is not odd,
$\lnq(1/p_i)=\tfrac{p_i^{\,q-1}-1}{1-q}\neq-\lnq p_i=\tfrac{1-p_i^{\,1-q}}{1-q}$,
so the two equivalent forms of $\Sq$ in Eq.~\eqref{eq:tsallis-entropy} are linked
by the escort reweighting $p_i^{\,q}\lnq p_i=-p_i\lnq(1/p_i)$ rather than by
oddness. The entropy must therefore retain the $p_i^{\,q}$ escort weight, whereas
the loss keeps the linear label weight $y_i$.

\paragraph{Related robust losses.} The $\mathcal{L}_q$ robust classification loss
and the generalized cross-entropy for learning with noisy
labels~\citep{zhang2018gce} are members of this family, and the popular focal
loss can be read as a $q$-style reweighting that down-weights easy,
high-probability examples. The same deformation underlies the robust
\emph{bi-tempered} logistic loss and two-temperature logistic regression, which
introduce \emph{separate} entropic indices for the $q$-logarithm and the
$q$-exponential (softmax) and are derived explicitly from the Tsallis
divergence~\citep{amid2019twotemperature,amid2019bitempered}.

\paragraph{Entropic regularization.} On the regularization side, adding
$\lambda\,\Sq(p)$ to an objective (as in entmax) controls the sparsity of the
resulting distribution, while the Tsallis divergence, a $q$-deformed KL
divergence, provides a tunable proximity measure for variational objectives and
distillation.

\subsection{Optimization and metaheuristics}

Heavy-tailed $q$-Gaussian distributions are also useful as \emph{search
distributions}. \emph{Generalized simulated annealing}~\citep{tsallis1996gsa}
samples candidate moves from a $q$-Gaussian visiting distribution and accepts
them with a $q$-deformed criterion; the heavy tails permit occasional long
jumps that escape local minima far more effectively than Gaussian
(classical) or Cauchy (fast) annealing, and the method has been widely adopted
for molecular and hyperparameter optimization. The same idea recurs in
evolutionary strategies and particle-swarm methods that use $q$-Gaussian
mutation/velocity distributions to balance local refinement against global
exploration. At the level of gradient-based learning, the \emph{$q$-gradient} of
\citet{soterroni2012qgradient} replaces the ordinary partial derivative with
Jackson's $q$-derivative, giving a secant-like operator whose non-local steps
are less prone to stalling in flat regions. We flag this case because it is a
genuine false friend: the $q$ of the $q$-gradient is the deformation parameter
of quantum ($q$-)calculus, \emph{not} the Tsallis entropic index, and the two
coincide only in both reducing to the classical object as $q\to1$. The
$q$-Gaussian visiting distribution of generalized simulated annealing is
Tsallis-derived; the $q$-gradient is not. Conflating the two indices, which is
easy to do given the shared notation, would attribute to nonextensive
statistics a method that belongs to a separate deformation. Across these methods the
entropic index again behaves as an explore/exploit dial, now in the space of
optimization trajectories rather than probabilities.

\subsection{Uncertainty, calibration, and representation learning}

Two further threads deserve mention. First, generalized entropies provide
\emph{uncertainty and diversity measures}: $\Sq$ and the related Tsallis/Hill
diversity indices are used as acquisition functions in active learning and as
diversity regularizers, where $q$ tunes how strongly rare classes or modes are
emphasized. Second, in self-supervised and contrastive representation learning,
the temperature-scaled softmax at the core of the InfoNCE objective can be
replaced or augmented by $q$-deformed counterparts, changing how hard negatives
are weighted; heavy-tailed $q$-Gaussian similarity kernels likewise alter the
geometry of the learned embedding space. These uses are less mature than sparse
attention or max-entropy RL, but they fit the same template and represent
natural extensions.

\section{Cross-cutting Perspectives}
\label{sec:perspectives}

\subsection{Nonextensive signatures in learning dynamics}
\label{sec:dynamics}

The applications of Section~\ref{sec:applications} \emph{import} $q$-statistics
into AI by design. A complementary and, for a statistical-physics audience,
more striking observation is that nonextensive statistics appears in deep
learning \emph{unbidden}: several robust empirical regularities of trained
networks are heavy-tailed in exactly the way the $q$-CLT and superstatistics
predict.

\paragraph{Heavy-tailed weight spectra.} The empirical spectral densities of
the weight matrices of well-trained networks are not those of random Gaussian
matrices; instead they develop heavy, approximately power-law tails, a
phenomenon analyzed as ``heavy-tailed self-regularization'' through random
matrix theory~\citep{martin2021}. Power-law spectra are the hallmark of strong
correlations between the matrix elements, the regime in which the
ordinary CLT fails and the $q$-CLT applies. A $q$-Gaussian description of the
weight entries, with the spectral exponent fixed by $q$, is therefore the
natural nonextensive reading of this effect, and links generalization quality
to a measurable entropic index.

\paragraph{Heavy-tailed gradient noise.} Stochastic gradient descent has been
shown to experience gradient noise whose distribution is heavy-tailed rather
than Gaussian, with a finite tail index $\alpha<2$ indicating divergent
variance~\citep{simsekli2019}. A line of follow-up work has hardened this into
theory: the SGD iterates themselves converge to a heavy-tailed stationary
law~\citep{gurbuzbalaban2021heavytail}, multiplicative noise is identified as a
generic mechanism producing such tails~\citep{hodgkinson2021multiplicative}, and
the tail index has been tied to generalization and to network compressibility
through the Hausdorff dimension of the
trajectory~\citep{simsekli2020hausdorff,barsbey2021heavytails}. This places the
dynamics outside the basin of the classical CLT and inside that of the
$q$-CLT / L\'evy regime: the natural stationary and increment distributions are
$q$-Gaussians (Section~\ref{sec:overview}), and the escape from sharp minima is
governed by heavy-tailed, jump-like moves rather than Brownian diffusion.
Superstatistics (Section~\ref{sec:superstat}) gives a mechanism: the
per-mini-batch curvature and gradient scale fluctuate across the dataset, and
averaging local Gaussian fluctuations over those scale fluctuations yields
$q$-exponential statistics with $q-1$ set by their relative variance.

\paragraph{Why this matters.} Read together, these observations suggest that
nonextensivity is not only a modeling choice one can adopt but a structural
property of how large networks train and generalize. This has two consequences.
First, it offers a falsifiable language for phenomena that are currently
described case by case: fit $q$ (or the tail index $\alpha$) and test it
against the superstatistical prediction. Second, it argues that the
``$q$-dial'' methods are not foreign impositions but matched to the intrinsic
statistics of the systems they act on: a $q$-Gaussian prior over weights or a
$q$-deformed noise model in SGD analysis is using the distribution the data
already exhibit. We regard quantifying these signatures, and connecting the
fitted $q$ to generalization and optimization guarantees, as one of the most
promising bridges between nonextensive statistical mechanics and AI.

\subsection{Information geometry and $q$-exponential families}

Information geometry supplies the most general of these unifications. The
$q$-exponential family, consisting of distributions of the form
$p(x;\theta)=\expq(\theta^\top T(x)-A_q(\theta))$, forms a statistical
manifold with a natural dually-flat
geometry, in which the $q$-deformed (Tsallis/$\alpha$-) divergence plays the
role of the KL divergence and induces a generalized Fisher
metric~\citep{amari2016,naudts2011,amari2011qexpgeometry}; the construction
extends to general $\phi$-deformed exponential families with a matching
geometric duality~\citep{korbel2019phideformed}. Many of the methods above are,
from this viewpoint, the same construction performed on the $q$-deformed
manifold rather than the exponential one: entmax is a Bregman/Fenchel--Young
projection with respect to the Tsallis convex conjugate; $q$-Gaussian models are
maximum-entropy distributions on this manifold; Tsallis-regularized RL is policy
optimization in the dual geometry; and even clustering and density estimation
have been recast over the associated \emph{tempered exponential
measures}~\citep{amid2023clustering}. This geometric picture explains why the
applications share structure, and it provides the natural setting in which to
transfer results (convergence rates, convexity, estimator efficiency) from
the BGS case to the Tsallis case.

\subsection{The unifying ``$q$-dial'' pattern}

Stepping back, the applications of Section~\ref{sec:applications} instantiate
one idea: take a Boltzmann--Gibbs--Shannon object, deform it by $q$, and obtain
a continuum whose endpoints are familiar (dense softmax, Gaussian, Shannon
cross-entropy at $q=1$) and whose interior offers sparsity, heavy tails, or
robustness. This has three practical consequences. (i) \emph{Backward
compatibility}: every Tsallis method degenerates to its standard counterpart at
$q=1$, so adopting it cannot do worse than the baseline if $q$ is chosen or
learned well. (ii) \emph{A single, interpretable hyperparameter}: $q$ carries a
consistent meaning across all these settings, namely the weighting of rare
versus frequent events, so intuition transfers. (iii) \emph{Learnability}: because $\Sq$
and the $q$-deformed maps are smooth in $q$ for fixed support, $q$ can in
principle be optimized by gradient descent, as the adaptively-sparse
Transformers already demonstrate.

\subsection{Learning the entropic index $q$}
\label{sec:learnq}

The ``$q$-dial'' framing raises the question of how $q$ should be set. In most
of the literature it is a fixed hyperparameter, chosen by hand or by grid
search. But the optimal $q$ is rarely universal. It depends on the task, the
modality, the layer, the attention head, and even the individual input, so
treating it as a constant both costs accuracy and discards an interpretable
diagnostic. The natural alternative is to make $q$ \emph{learnable}, optimizing
it jointly with the model's other parameters rather than fixing it a priori.
This subsection collects the considerations that make learning $q$ both
attractive and delicate.

\paragraph{Differentiability and gradients.} For the objects whose dependence on
$q$ is explicit, namely the Tsallis-entropy regularizer and the $q$-Gaussian and
$q$-cross-entropy losses, the gradient with respect to $q$ is available in
closed form. For the entropy, differentiating Eq.~\eqref{eq:tsallis-entropy}
(with $k=1$) gives
\begin{equation}
  \frac{\partial \Sq(p)}{\partial q}
  \;=\;
  \frac{-(q-1)\sum_i p_i^{\,q}\ln p_i \;-\;\big(1-\sum_i p_i^{\,q}\big)}
       {(q-1)^2},
  \label{eq:dSdq}
\end{equation}
which is smooth for all $q\neq1$ and has a finite limit as $q\to1$, so $q$ can be
carried as an ordinary differentiable parameter. The subtler case is the
\emph{argmax-defined} maps such as $\alpha$-entmax
(Eq.~\eqref{eq:entmax}), where the output $p^\star(q)$ has no explicit form. Here
one differentiates through the optimality (KKT) conditions: on the support
$\mathcal{S}$ of the solution, stationarity reads
$z_i + \partial_{p_i}\Sq(p^\star) = \tau$ for $i\in\mathcal{S}$, and applying the
implicit function theorem to this system yields $\partial p^\star/\partial q$ as
the solution of a small linear system over $\mathcal{S}$. This is exactly the
Jacobian that the entmax backward pass computes, so learning $q$ requires no new
machinery, only that the gradient be propagated to the index as well as to the
logits. The one genuine non-smoothness is the Tsallis cut-off: as $q$ varies, an
element can enter or leave the support, and at those (measure-zero) events
$p^\star(q)$ is continuous but not differentiable. One-sided derivatives or a
small amount of smoothing handle this in practice.

\paragraph{Where and how $q$ lives.} Learnability invites a spectrum of
parameterizations of increasing expressiveness:
\begin{itemize}[nosep]
  \item \textbf{Global.} A single scalar $q$ for the whole model, the cheapest
        upgrade over a fixed hyperparameter, learned by gradient descent or
        optimized on a validation set.
  \item \textbf{Structured.} A separate $q$ per layer, per attention head, or per
        channel. The adaptively sparse Transformers
        of~\citet{correia2019adaptively} are the canonical instance: each head
        learns its own $\alpha=q$, and the trained values reveal that some heads
        specialize to dense, broadly-attending behavior ($q\!\approx\!1$) while
        others become highly sparse ($q\!>\!1$). The learned index thus doubles
        as an interpretability signal.
  \item \textbf{Conditional (amortized).} An input-dependent index
        $q = g_\phi(x)$ produced by a small network, an ``entropic gate'' that
        decides, per token or per state, how sparse or heavy-tailed to be. This
        is the least explored of the three: it lets a model be dense where
        context is ambiguous and sparse where it is decisive.
\end{itemize}
To keep $q$ in a valid and well-behaved range it is reparameterized through an
unconstrained variable $\theta$: $q = 1 + \mathrm{softplus}(\theta)$ confines it
to the sparsity regime $q>1$, while a scaled sigmoid $q \in (1,3)$ additionally
guarantees a normalizable $q$-Gaussian (cf.\ the moment conditions in
Section~\ref{sec:overview}).

\paragraph{Optimization strategies and failure modes.} Jointly minimizing the
training loss over $q$ is the simplest recipe but is prone to degeneracy,
because the model can often reduce \emph{training} loss by moving $q$ to where
the regularizer stops binding, for example collapsing entmax back toward
softmax, or driving a robust $q$-loss to downweight hard examples. Three remedies
recur: (i) optimize $q$ as a \emph{hyperparameter} on held-out data via bilevel /
hypergradient methods, decoupling it from the objective it is meant to shape;
(ii) place a \emph{prior} on $q$ and learn its posterior, the Bayesian view, of
which the long-standing practice of learning the degrees of freedom of a
Student-$t$ model is a special case ($q$ and the degrees of freedom are in
one-to-one correspondence); and (iii) \emph{schedule} $q$, annealing from
$q\!\approx\!1$ (dense, stable, well-conditioned gradients early in training)
toward larger $q$ (sparser, more committed) as training proceeds. A final caveat
is \emph{confounding with temperature}: both $q$ and a softmax/entropy
temperature $\tau$ control peakedness, so learning them simultaneously is
ill-posed unless one is fixed or they are explicitly coupled. In the
reinforcement-learning setting this confounding is acute, since an unconstrained
joint update on $q$ and the entropy coefficient can silently collapse
exploration. \citet{vasylenko2026longcontext} address this directly in the
attention setting by learning the index and the temperature together under an
explicit length-dependent coupling, which is the kind of constraint that makes
the joint problem well posed.

\paragraph{Summary.} A learnable $q$ turns nonextensivity from a modeling
assumption into a quantity estimated from data. It costs one extra parameter
and a known gradient, and it yields an
interpretable readout of how much sparsity or how heavy a tail each part of the
network requires. The open work is less about feasibility than about doing
it well at scale: input-conditional indices, principled (validation- or
prior-based) objectives that avoid the degeneracies above, and theory for the
resulting joint optimization. We return to this in Section~\ref{sec:future}.

\paragraph{An empirical illustration.} To make the discussion concrete, rather
than to benchmark it, we run two small, fully reproducible experiments in
which $q$ is \emph{learned by gradient descent} and compared against the
Boltzmann--Gibbs--Shannon ($q=1$) baseline; the complete JAX code is in the
repository (\texttt{experiments/learnable\_q.py}, fixed seed). These are
minimal proofs of concept, not evidence that the idea scales: for that,
\citet{vasylenko2026longcontext} already demonstrate a learnable entropic
index on real language models at the scale that matters, and we discuss their
result alongside ours below. Listing~\ref{lst:learnq}
shows the shared core: the Tsallis primitives, the range-constraining
reparameterizations introduced above, and the property the construction rests
on, namely that $q$ is an ordinary differentiable parameter, so
\lstinline{jax.value_and_grad} propagates a gradient into it alongside the model
weights.

\begin{lstlisting}[style=qjpy, caption={Shared core: $q$ is a differentiable
parameter, kept in range by a transform and learned with the same autodiff
update as the weights.}, label={lst:learnq}, float=t]
import jax, jax.numpy as jnp
from jax.scipy.special import gammaln

def ln_q(x, q):                          # Tsallis q-logarithm, ln_1 = log
    omq = 1.0 - q
    return jnp.where(jnp.abs(omq) < 1e-6, jnp.log(x), (x**omq - 1.0) / omq)

def student_t_logpdf(x, loc, log_scale, q):     # normalized q-Gaussian, 1<q<3
    a, nu = jnp.exp(log_scale), (3.0 - q) / (q - 1.0)
    z = (x - loc) / a
    return (gammaln((nu + 1)/2) - gammaln(nu/2) - 0.5*jnp.log(nu*jnp.pi)
            - log_scale - 0.5*(nu + 1)*jnp.log1p(z*z / nu))

def q_cross_entropy(logits, y, q):              # -sum_i y_i ln_q(p_i)
    p = jnp.exp(jax.nn.log_softmax(logits))     # (clamped in the released code)
    return -jnp.sum(y * ln_q(p, q), axis=-1)

q_A = lambda t: 1.0 + 2.0*jax.nn.sigmoid(t)     # (1,3): heavy-tailed q-Gaussian
q_B = lambda t: jax.nn.sigmoid(t)               # (0,1): robust, bounded q-loss

loss = lambda p: -jnp.mean(student_t_logpdf(x, p["loc"], p["log_scale"],
                                            q_A(p["theta"])))
value, grads = jax.value_and_grad(loss)(params) # gradient flows into theta = q
params = adam_step(params, grads)               # q is learned like any weight
\end{lstlisting}

\begin{figure}[t]
\centering
\includegraphics[width=\textwidth]{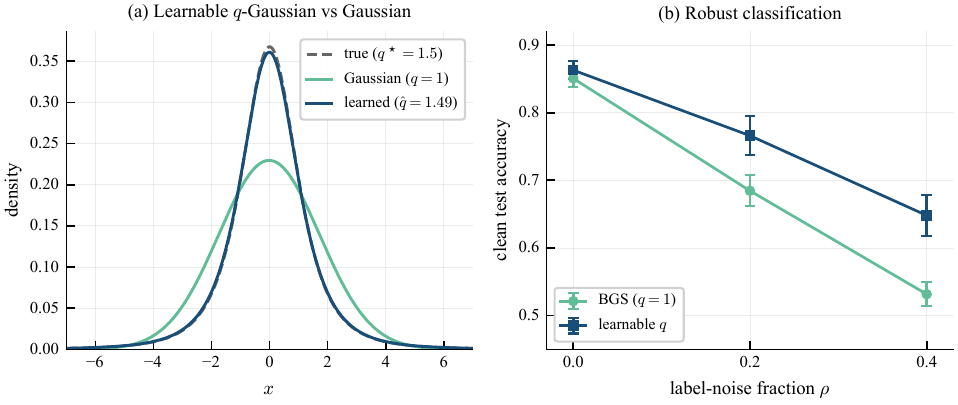}
\caption{Empirical illustration of a learnable entropic index $q$ against the
$q=1$ baseline (code: \texttt{experiments/learnable\_q.py}).
\textbf{(a)} Densities fitted by maximum likelihood to a heavy-tailed sample:
the Gaussian ($q{=}1$) must over-broaden to cover the tails, while the
learnable $q$-Gaussian recovers the true peak and tail ($\hat q=1.49$).
\textbf{(b)} Clean-test accuracy of an MLP trained with $q$-cross-entropy under
label noise (mean over $8$ seeds; error bars are $\pm1$ s.d.): learning $q<1$ on
held-out clean data resists the label-noise memorization that degrades the
$q{=}1$ baseline, with the gap widening as the noise level grows.}
\label{fig:learnq}
\end{figure}

In the first experiment (Table~\ref{tab:learnq}a, Fig.~\ref{fig:learnq}a) we draw
data from a heavy-tailed $q$-Gaussian ($q^\star=1.5$, i.e.\ a Student-$t$ with
three degrees of freedom) and fit two density models by maximum likelihood: a
Gaussian (the $q=1$ model) and a $q$-Gaussian whose index is learned through
$q = 1 + 2\,\sigma(\theta)\in(1,3)$. The learnable model recovers $\hat q=1.49$
and the true scale ($1.02$ against the generative $1.0$), whereas the Gaussian
can fit the tails only by inflating its scale to $1.74$; the learned-$q$ model
improves the held-out log-likelihood by $0.14$ nat per sample.

In the second experiment (Table~\ref{tab:learnq}b, Fig.~\ref{fig:learnq}b) a
small multilayer perceptron is trained on a three-class problem with a fraction
$\rho$ of the training labels randomly corrupted, using the $q$-cross-entropy
$\mathcal{L}_q=-\sum_i y_i\lnq \hat p_i$. Here the robust, \emph{bounded} regime
is $q<1$, since the $q$-logarithm caps the penalty on confidently-wrong
predictions, whereas the $q>1$ side is more aggressive than ordinary
log-loss~\citep{zhang2018gce}, so we learn $q=\sigma(\theta)\in(0,1)$. Because jointly minimizing the
\emph{noisy training} loss over $q$ is degenerate (it drives $q$ toward whatever
fits the corrupted labels), we apply remedies (i)--(ii) above: $q$ is selected on
a small held-out \emph{clean} validation set under a weak Gaussian prior centered
at the BGS value $q=1$. Averaged over eight independent seeds, the learned index
decreases monotonically with the noise level ($\hat q\approx0.33,0.17,0.08$ at
$\rho=0,0.2,0.4$), staying well below $1$, and the resulting classifier retains
clean-test accuracy that the $q=1$ baseline loses by memorizing the corrupted
labels: $+8.1$ and $+11.6$ percentage points at $\rho=0.2$ and
$\rho=0.4$ respectively (Table~\ref{tab:learnq}b).

One feature of Table~\ref{tab:learnq}b deserves comment, since it runs against
the naive expectation. At $\rho=0$ the selected index is $\hat q\approx0.33$
rather than the $\hat q\to1$ one might anticipate from the absence of label
noise. Two things are at work. First, the classes in this task overlap by
construction, so even with uncorrupted labels a fraction of training points are
not separable and behave like noise; a bounded loss is mildly preferable on
them, and the clean-validation criterion duly selects $q<1$. Second, and more
importantly, the objective is nearly flat in $q$ over much of $(0,1)$ once the
data are clean, so the selected value is weakly identified: the accuracy
difference against the baseline at $\rho=0$ is $1.2$ points with overlapping
standard deviations, which is not a meaningful gap. The honest reading is that
$q$ is well determined only when there is contamination for it to respond to,
which is visible in the standard deviations shrinking from $0.06$ to $0.02$ as
$\rho$ rises from $0$ to $0.2$. This is a limitation of using the learned index
as a diagnostic: it is informative when the deformation is doing work, and
close to arbitrary when it is not.

These experiments are deliberately small and illustrative: their point is only
that a single extra scalar, learned with the existing autodiff machinery, lets
a model choose its own degree of nonextensivity, with $q=1$ always available as
a fallback. The claim that this scales is not ours to make on this evidence; it
is made, convincingly, by \citet{vasylenko2026longcontext}, whose jointly
learned index and temperature improve long-context generalization on real
language models. We see the two results as complementary rather than
overlapping: theirs shows the mechanism works at the scale that matters, ours
isolates it on toy problems small enough to inspect $\hat q$ directly and check
it against a known ground truth.

\begin{table}[t]
\centering\small
\caption{Learnable $q$ versus the Boltzmann--Gibbs--Shannon ($q=1$) baseline.
\textbf{(a)} Maximum-likelihood density estimation on a heavy-tailed sample
($q^\star=1.5$, scale $1$): the learnable $q$-Gaussian recovers the index and
scale and improves held-out likelihood. \textbf{(b)} Clean-test accuracy of an
MLP trained with $q$-cross-entropy under label noise $\rho$; the learned $q$
(selected on a clean validation set) falls below $1$ and improves robustness.}
\label{tab:learnq}
\renewcommand{\arraystretch}{1.15}
\textbf{(a) Heavy-tailed density estimation}\par\vspace{1pt}
\begin{tabular}{lccc}
\toprule
Model & $q$ & fitted scale & test log-lik.\ (nat) \\
\midrule
Gaussian (BGS, $q{=}1$)  & $1.00$ (fixed) & $1.74$ & $-1.936$ \\
$q$-Gaussian (learned)   & $1.49$         & $1.02$ & $\mathbf{-1.799}$ \\
\addlinespace[1pt]
\textit{generative truth} & $1.50$        & $1.00$ & n/a \\
\bottomrule
\end{tabular}
\par\vspace{6pt}
\textbf{(b) Classification under label noise}
(clean-test accuracy, mean\,$\pm$\,s.d.\ over $8$ seeds)\par\vspace{1pt}
\begin{tabular}{cccc}
\toprule
noise $\rho$ & BGS ($q{=}1$) & learnable $q$ & learned $q$ \\
\midrule
$0.0$ & $0.851 \pm 0.013$ & $0.863 \pm 0.013$          & $0.33 \pm 0.06$ \\
$0.2$ & $0.685 \pm 0.023$ & $\mathbf{0.766 \pm 0.029}$ & $0.17 \pm 0.02$ \\
$0.4$ & $0.532 \pm 0.018$ & $\mathbf{0.648 \pm 0.031}$ & $0.08 \pm 0.04$ \\
\bottomrule
\end{tabular}
\end{table}

\subsection{Pitfalls and caveats}

Several recurring pitfalls deserve explicit warning:
\begin{itemize}[nosep]
  \item \textbf{Choice of constraints.} Maximum-$\Sq$ inference depends on how
        expectation constraints are defined (ordinary versus normalized
        $q$-expectations / escort distributions). Different conventions give
        different distributions; results must state which is used.
  \item \textbf{Normalizability and moments.} $q$-Gaussians are only
        normalizable for $q<3$ and have finite variance only for $q<5/3$.
        Naively plugging large $q$ into a model that assumes finite second
        moments is a frequent error.
  \item \textbf{Identifiability and overlap.} Because $q$-Gaussians coincide
        with Student-$t$ and the entmax/sparse-RL constructions coincide with
        previously known projections, ``Tsallis'' results sometimes restate
        existing methods. The value added is the unifying parameterization and
        the principled extrapolation it enables, not novelty for its own sake.
  \item \textbf{Optimization of $q$.} The cut-off in $\expq$ introduces
        non-smoothness at the support boundary, so jointly learning $q$ requires
        care (e.g.\ smoothing, or restricting to regimes where the support is
        stable).
  \item \textbf{Empirical justification.} Heavy tails should be demonstrated,
        not assumed. Fitting a $q$ and reporting that it differs from $1$ is
        evidence for nonextensive structure only when accompanied by proper
        model comparison.
\end{itemize}

\section{Open Challenges and Future Directions}
\label{sec:future}

We highlight directions where the Tsallis perspective seems most likely to pay
off.

\subsection{$q$ as a learned, context-dependent inductive bias}

Adaptively-sparse Transformers learn one $\alpha$ per head; the natural
generalization (Section~\ref{sec:learnq}) is to let $q$ depend on the input, the
layer, the token, or the RL state, turning the entropic index into a learned
gating signal for sparsity and tail behavior. The outstanding problems are
amortized, input-conditional indices at scale and objectives (validation-based
or Bayesian) that avoid the degenerate joint optima discussed above.

\subsection{Foundation-model-scale evaluation}

Efficient kernels~\citep{goncalves2025adasplash} and length-adaptive
formulations~\citep{vasylenko2026longcontext} have brought $\alpha$-entmax
attention to encoder models and mid-size decoders, but $q$-deformed losses,
priors, and policies remain validated mostly at modest scale. Whether the
efficiency, interpretability, and robustness benefits persist or compound at
frontier model scale, and whether a learned $q$ stays stable under the
optimization regimes used there, is an open and consequential empirical
question.

\subsection{Heavy-tailed training dynamics}

The observed heavy-tailed gradient noise and weight spectra in deep networks
invite a $q$-statistical treatment of optimization and generalization, including
$q$-Gaussian noise models in SGD analysis and $q$-CLT-based accounts of why
heavy tails arise.

\subsection{Robustness, safety, and calibration}

$q$-deformed losses and priors give a principled dial for robustness to label
noise, distribution shift, and outliers; connecting the entropic index to formal
robustness or calibration guarantees would make it usable as a tool for
trustworthy AI.

\subsection{Unified software and theory}

A coherent library exposing $q$ across attention, losses, priors, and RL,
together with convergence and generalization theory transferred through the
information-geometric picture, would lower the barrier to adoption and let
practitioners treat nonextensivity as a routine modeling choice.

\subsection{Diffusion and generative modeling with heavy tails}

Systematic study of $q$-Gaussian / $\alpha$-stable forward processes in
diffusion models, and of $q$-deformed likelihoods in autoregressive generation,
is largely unexplored territory with clear motivation from the heavy-tailed
nature of real data.

\section{Conclusion}
\label{sec:conclusion}

Tsallis statistics extends the Boltzmann--Gibbs--Shannon framework that
underpins machine learning through a single entropic index $q$, supplying a
self-consistent algebra ($q$-entropy, $q$-exponential and $q$-logarithm,
$q$-central limit theorem, and $q$-Gaussian) that reduces to the familiar
toolkit at $q=1$. Surveying its appearances across AI, we find that a wide
range of methods, including sparse attention ($\alpha$-entmax, sparsemax),
maximum-entropy and sparse reinforcement learning, heavy-tailed generative and
probabilistic models, robust loss functions, and heavy-tailed optimization,
instantiate the same ``$q$-dial'' design pattern: a tunable, backward-compatible
interpolation between dense/light-tailed and sparse/heavy-tailed behavior. On
this reading, $q$ is not a fixed physical constant but a general-purpose,
interpretable, and in principle learnable inductive bias. The
information-geometric picture of $q$-exponential families explains why these methods share
structure and points the way to transferring theory between them. We expect the
most impactful near-term progress to come from treating $q$ as a learned,
context-dependent parameter, from rigorous large-scale evaluation, and from a
$q$-statistical understanding of the heavy-tailed dynamics already pervasive in
deep learning.

\section*{CRediT authorship contribution statement}

\textbf{Kleyton da Costa:} Conceptualization, Methodology, Software,
Investigation, Visualization, Writing of the original draft, Review and
editing. \textbf{Bernardo Modenesi:} Conceptualization, Methodology,
Validation, Supervision, Review and editing.

\section*{Declaration of competing interest}

The authors declare that they have no known competing financial interests or
personal relationships that could have appeared to influence the work reported
in this paper.

\section*{Data availability}

No new data were generated in this study. The source code reproducing all
experiments and figures, together with the \texttt{qjax} library described in
Appendix~\ref{app:qjax}, is publicly available under the MIT license at
\url{https://github.com/Kleyt0n/qjax}.

\section*{Declaration of generative AI in the writing process}

The authors used a generative AI assistant to help with language editing,
literature cross-checking, and consistency checks on the manuscript. The
authors reviewed and edited all content and take full responsibility for the
substance and accuracy of the published article.

\section*{Acknowledgments}

The authors would like to thank Professor Constantino Tsallis for his insightful feedback on an early draft of this paper, and for his pioneering work in nonextensive statistical mechanics that inspired it. Any remaining errors or omissions are our own.

\bibliographystyle{elsarticle-num-names}
\bibliography{references}

\appendix
\sloppy

\section{\texttt{qjax}: Tsallis Statistics for Artificial Intelligence in JAX}
\label{app:qjax}

The constructions reviewed in this paper are released as \texttt{qjax}, an
open-source (MIT-licensed) companion library that exposes the $q$-deformed
primitives of Tsallis statistics as pure, differentiable,
\texttt{jit}/\texttt{vmap}-friendly JAX functions. The library is organized
around three design principles: \emph{purity} (every function is side-effect
free and composes with JAX transformations); \emph{the $q\to1$ limit} (every
primitive provably recovers its Boltzmann--Gibbs--Shannon counterpart as
$q\to1$); and \emph{finite gradients} (all functions remain differentiable with
finite gradients everywhere, including exactly at $q=1$). Because the
entropic index $q$ is just another function argument, it can be held fixed
\emph{or} learned end-to-end by gradient descent, which is the central thesis of
this paper realized in code. \texttt{qjax} requires Python~$\geq3.10$ and
JAX~$\geq0.4.30$ and runs on any JAX backend (CPU, GPU, or TPU).

\subsection{Building blocks}
\label{app:qjax-blocks}

Every primitive is a single closed form in $q$ and reduces to its classical
counterpart as $q\to1$ (Table~\ref{tab:qjax-api}). All functions are top-level
exports of the \texttt{qjax} package.

\begin{table}[h]
\centering
\footnotesize
\setlength{\tabcolsep}{2pt}
\renewcommand{\arraystretch}{1.5}
\caption{Core \texttt{qjax} primitives, their definitions, and the
Boltzmann--Gibbs--Shannon object each recovers as $q\to1$. Here
$[\,\cdot\,]_+=\max(\cdot,0)$ is the Tsallis cut-off, $C_q$ the $q$-Gaussian
normalization, and $\Delta$ the probability simplex; \texttt{tsallis\_entmax}
is exactly sparsemax at $q=2$.}
\label{tab:qjax-api}
\begin{tabular}{@{}lll@{}}
\toprule
\textbf{\texttt{qjax} function} & \textbf{definition} & \textbf{limit }$q\to1$\\
\midrule
\texttt{q\_log(x, q)} & $\lnq x=\dfrac{x^{1-q}-1}{1-q}$ & $\ln x$\\
\texttt{q\_exp(x, q)} & $\expq x=[1+(1-q)x]_+^{1/(1-q)}$ & $e^{x}$\\
\texttt{tsallis\_entropy(p, q)} & $\Sq(p)=\dfrac{1-\sum_i p_i^{\,q}}{q-1}$
  & $-\sum_i p_i\ln p_i$\\
\texttt{tsallis\_cross\_entropy(p, y, q)} & $H_q(y,p)=-\sum_i y_i\lnq p_i$
  & $-\sum_i y_i\ln p_i$\\
\texttt{tsallis\_divergence(p, r, q)}
  & $D_q(p\Vert r)=\dfrac{\sum_i p_i^{\,q}r_i^{\,1-q}-1}{q-1}$
  & $\mathrm{KL}(p\Vert r)$\\
\texttt{q\_gaussian\_pdf(x, q, beta)}
  & $\mathcal{G}_q(x)=\dfrac{\sqrt{\beta}}{C_q}\,\expq(-\beta x^2)$
  & $\sqrt{\beta/\pi}\,e^{-\beta x^2}$\\
\texttt{tsallis\_entmax(z, q)}
  & $\operatorname{entmax}_q(z)=\argmax_{p\in\Delta}\langle p,z\rangle+\Sq(p)$
  & $\softmax(z)$\\
\bottomrule
\end{tabular}
\end{table}

The accompanying \texttt{q\_add}, \texttt{q\_diff}, \texttt{q\_prod}, and
\texttt{q\_div} implement the $q$-algebra, \texttt{sample} draws from the
$q$-Gaussian, and \texttt{qjax.plots} provides a small plotting layer, built on
a sequential green--blue ramp in which colour encodes the ordering in $q$, used
to produce every figure in this paper.

\subsection{Installation and quickstart}
\label{app:qjax-install}

The package is installed from PyPI with either \texttt{pip} or \texttt{uv}:

\begin{lstlisting}[style=qjpy, numbers=none, caption={Installing \texttt{qjax}
from PyPI.}, label={lst:qjax-install}]
pip install qjax        # or:  uv add qjax
\end{lstlisting}

\noindent A first session exercises the whole surface (deformed functions,
information measures, the $q$-Gaussian, and sparse activations), each
reducing to its classical analogue at $q=1$:

\begin{lstlisting}[style=qjpy, caption={Quickstart: the core \texttt{qjax}
primitives. Every call recovers its Boltzmann--Gibbs--Shannon counterpart as
$q\to1$.}, label={lst:qjax-quickstart}]
import jax, jax.numpy as jnp
import qjax

# q-deformed functions (recover log / exp as q -> 1)
qjax.q_log(2.0, q=1.5)
qjax.q_exp(1.0, q=1.5)

# Tsallis information measures
p = jnp.array([0.5, 0.3, 0.2])
qjax.tsallis_entropy(p, q=2.0)         # -> Shannon entropy as q -> 1
qjax.tsallis_divergence(p, p, q=2.0)   # -> KL divergence as q -> 1

# q-Gaussian distribution (heavy-tailed for 1 < q < 3)
x = jnp.linspace(-4, 4, 100)
qjax.q_gaussian_pdf(x, q=1.5, beta=1.0)
samples = qjax.sample(jax.random.PRNGKey(0), q=1.5, beta=1.0, shape=(1000,))

# Sparse softmax: q=1 -> softmax, q=2 -> sparsemax (exact zeros)
qjax.tsallis_entmax(jnp.array([2.0, 1.0, -1.0]), q=2.0)
\end{lstlisting}

\subsection{Examples}
\label{app:qjax-examples}

The examples below instantiate the single ``$q$-dial'' principle across several
learning tasks. For each we state the governing equation and the corresponding
\texttt{qjax} core; throughout, $q$ may be held fixed or treated as a trainable
parameter, constrained to its admissible range by a smooth reparameterization
$q = q_{\min} + (q_{\max}-q_{\min})\,\sigma(\theta)$ and optimized jointly with
the model.

\subsubsection{Sparse attention}
\label{app:ex-attention}

Tsallis-entmax replaces the softmax attention map. For scores
$s_j = \langle \mathbf{q}_{\text{qry}}, \mathbf{k}_j\rangle/\sqrt{d}$ over
positions $j$, the attention weights and pooled context are
\begin{equation}
  \alpha_j = \big[(q-1)(s_j-\tau)\big]_+^{\,1/(q-1)},
  \qquad
  \mathbf{c} = \sum_j \alpha_j\,\mathbf{v}_j,
  \label{eq:entmax-attn}
\end{equation}
where the threshold $\tau$ is set so that $\sum_j \alpha_j = 1$. At $q=1$ this is
dense $\softmax$ attention; for $q>1$ every score below $\tau$ is clipped to
\emph{exactly} zero, so pure-distractor positions receive no weight, and a
learnable $q$ lets gradient descent select the attention sparsity that best fits
the task.

\begin{lstlisting}[style=qjpy, numbers=none, float=t, caption={Sparse attention with
\texttt{tsallis\_entmax} (Eq.~\eqref{eq:entmax-attn}).}, label={lst:qjax-attn}]
import jax.numpy as jnp
import qjax

def forward(params, x, q):
    scores  = (x @ params["w_key"]) @ params["query"] / jnp.sqrt(D_MODEL)
    attn    = qjax.tsallis_entmax(scores, q=q, axis=-1, num_iters=25)
    context = jnp.einsum("nl,nld->nd", attn, x @ params["w_val"])
    return context @ params["w_out"] + params["b_out"], attn
\end{lstlisting}

\begin{figure}[t]
\centering
\includegraphics[width=\textwidth]{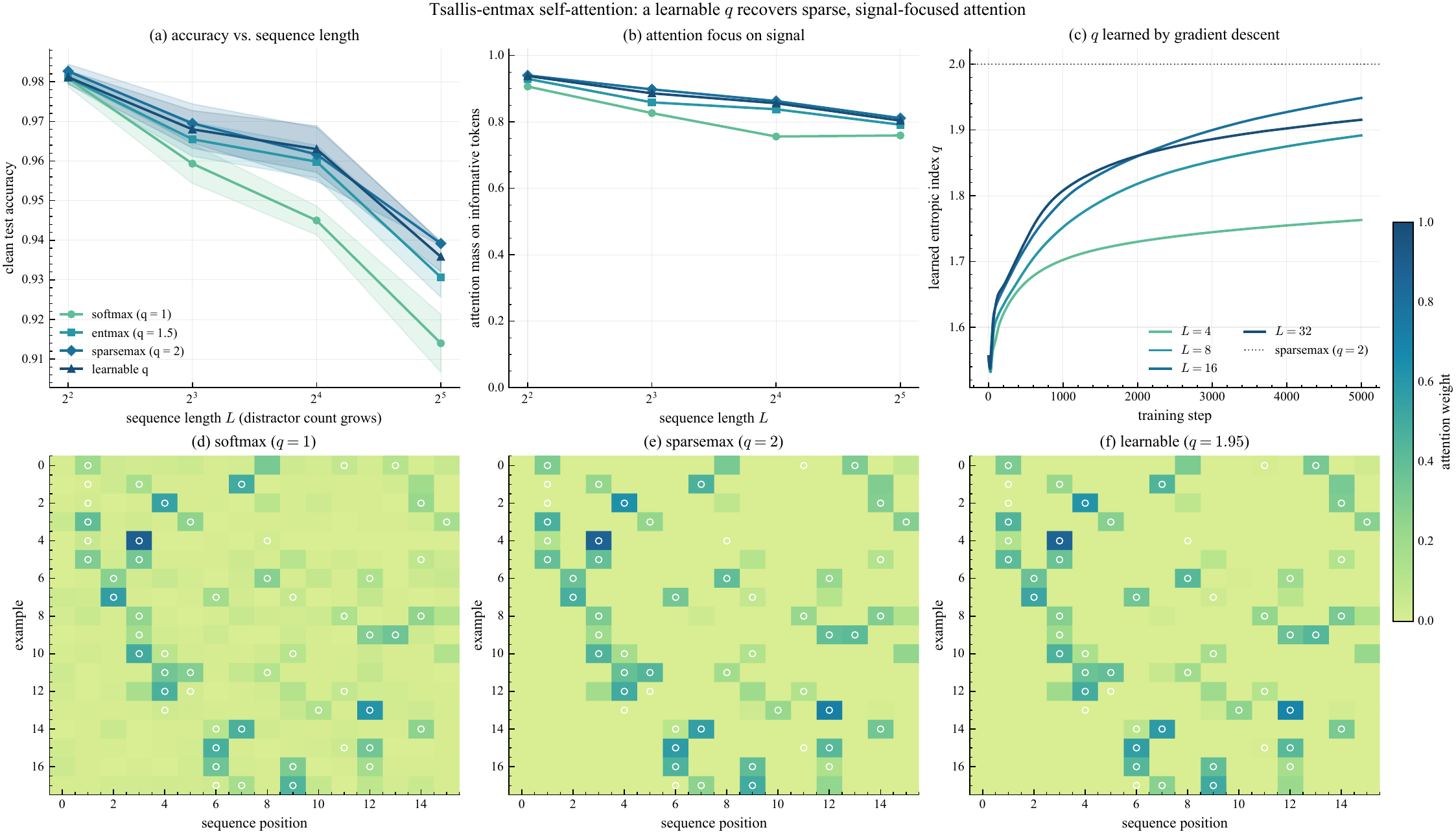}
\caption{Sparse attention on a needle-in-a-haystack retrieval task
(Eq.~\eqref{eq:entmax-attn}). \textbf{(a)} Test accuracy versus sequence length
$L$ (log scale): as distractors accumulate, dense $\softmax$ ($q=1$) degrades
while $\operatorname{entmax}$ ($q=1.5$), sparsemax ($q=2$), and the learnable-$q$
model stay robust. \textbf{(b)} Attention mass placed on the truly informative
tokens, where the sparse variants concentrate weight on signal. \textbf{(c)} The
entropic index $q$ learned by gradient descent over training, converging toward
the sparsemax regime ($q\to2$) across all sequence lengths. \textbf{(d--f)}
Learned attention maps (examples $\times$ positions; white circles mark the true
informative tokens) for $\softmax$ ($q=1$, diffuse), sparsemax ($q=2$, sharp),
and learnable $q$ (sharp, with $q$ discovered end-to-end).}
\label{fig:app-attention}
\end{figure}

\subsubsection{Robust classification}
\label{app:ex-classification}

A classifier is trained under label noise with the Tsallis cross-entropy of the
softmax output $p$ against the one-hot target with true class $c$,
\begin{equation}
  H_q(y,p) = -\lnq p_c = \frac{1 - p_c^{\,1-q}}{1-q}
  \;\xrightarrow{\,q\to1\,}\; -\ln p_c .
  \label{eq:tsallis-ce}
\end{equation}
For $q=1$ this is the ordinary (unbounded) cross-entropy, so a confidently
mislabeled point produces an arbitrarily large gradient and the network
memorizes the noise. For $q<1$ the loss is \emph{bounded} by $1/(1-q)$; its
gradient saturates on hard or corrupted examples, and clean-set accuracy
degrades far more gracefully. Making $q$ learnable lets the model descend to the
robust regime on its own.

\begin{lstlisting}[style=qjpy, numbers=none, float=t, caption={Label-noise-robust
training with the Tsallis cross-entropy loss
(Eq.~\eqref{eq:tsallis-ce}).}, label={lst:qjax-clf}]
import jax.numpy as jnp
import qjax

def loss_fn(params, x, y_onehot, q):
    p = jnp.clip(jax.nn.softmax(logits(params, x), axis=-1), 1e-7, 1.0)
    return jnp.mean(qjax.tsallis_cross_entropy(p, y_onehot, q=q, axis=-1))
\end{lstlisting}

\begin{figure}[t]
\centering
\includegraphics[width=\textwidth]{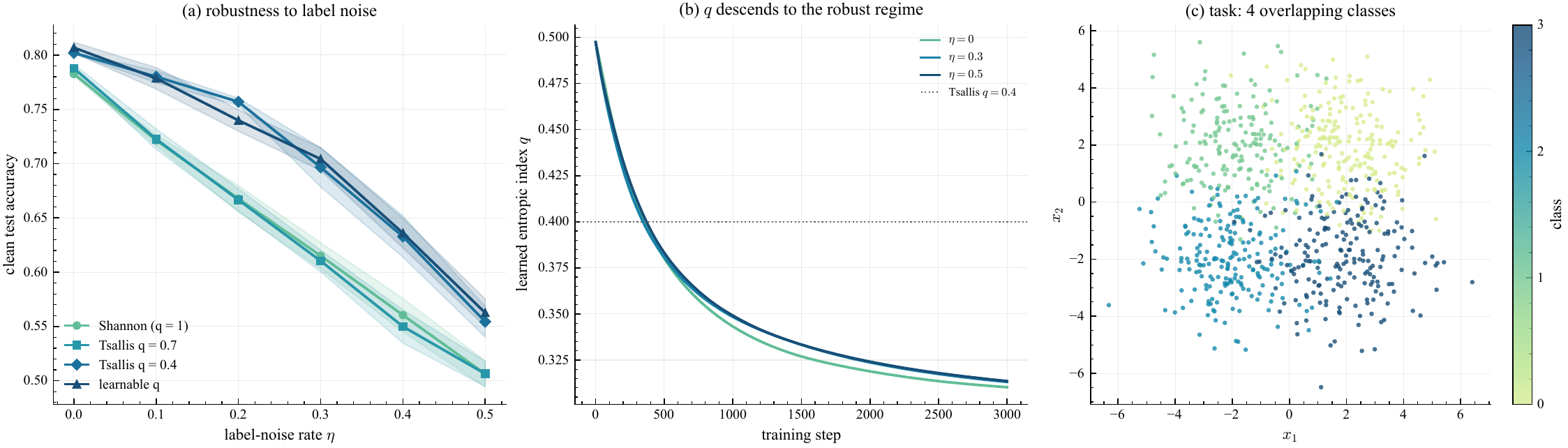}
\caption{Label-noise-robust classification with the Tsallis cross-entropy
(Eq.~\eqref{eq:tsallis-ce}). \textbf{(a)} Clean-test accuracy versus label-noise
rate $\eta$: the bounded Tsallis losses ($q=0.7,\,0.4$) and the learnable-$q$
model degrade far more gracefully than ordinary cross-entropy (Shannon, $q=1$).
\textbf{(b)} Starting from $q=0.5$, the learned index descends during training
into the robust regime ($q\approx0.31$) at every noise level shown
($\eta=0,\,0.3,\,0.5$), discovered by gradient descent. \textbf{(c)} The
synthetic two-dimensional task (four overlapping Gaussian classes).}
\label{fig:app-classification}
\end{figure}

\begin{figure}[t]
\centering
\includegraphics[width=\textwidth]{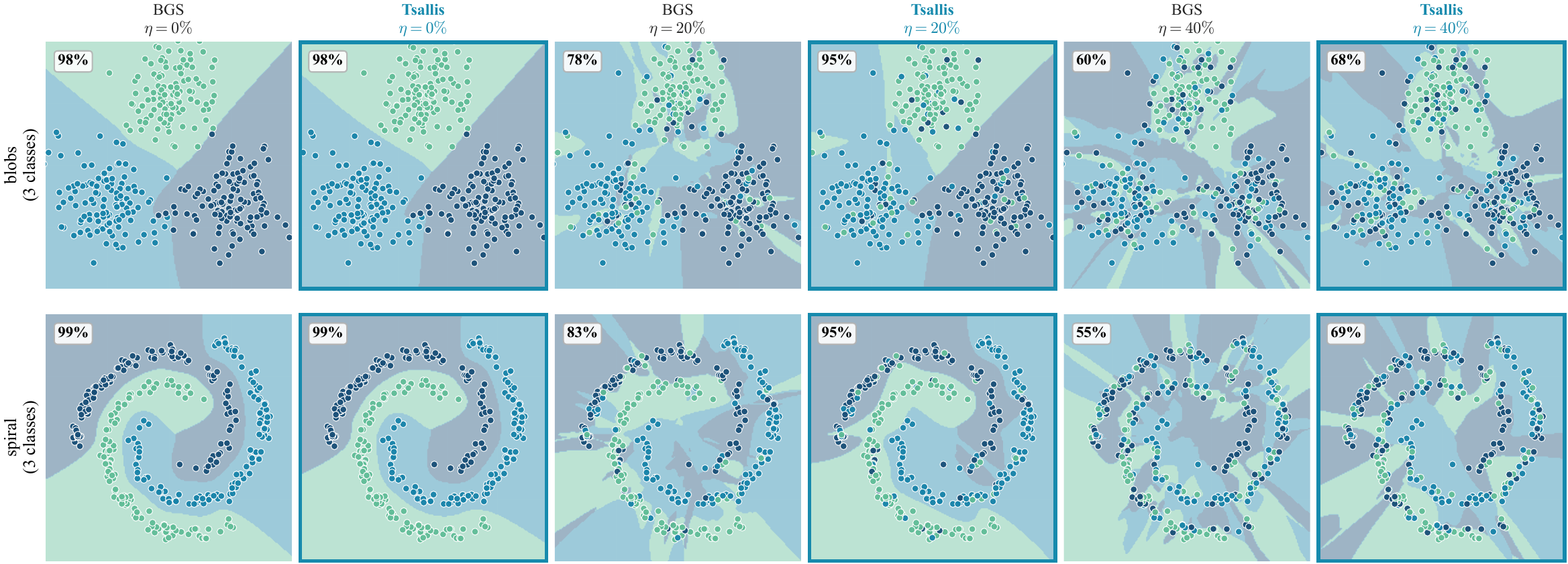}
\caption{Decision boundaries under label noise for two synthetic geometries
(blobs and spirals) at noise levels $\eta=0,\,0.2,\,0.4$. Ordinary cross-entropy
(Shannon, $q=1$) carves spurious islands around mislabeled points, whereas the
bounded Tsallis loss ($q=0.3$) keeps clean, smooth boundaries; the test accuracy
is annotated in each panel.}
\label{fig:app-classification-boundaries}
\end{figure}

\subsubsection{Node classification on graphs}
\label{app:ex-nodeclf}

The same bounded loss transplants onto a graph unchanged. A two-layer graph
convolutional network propagates features $X$ over the symmetrically normalized
adjacency $\hat A = \tilde D^{-1/2}(A+I)\,\tilde D^{-1/2}$ and is trained on the
labeled nodes $\mathcal{T}$ with the Tsallis cross-entropy of
Eq.~\eqref{eq:tsallis-ce},
\begin{equation}
  Z = \hat A\,\relu\!\big(\hat A X W^{(1)}\big)W^{(2)},
  \qquad
  \mathcal{L}_q = \frac{1}{|\mathcal{T}|}\sum_{i\in\mathcal{T}}
    \big(-\lnq p_{i,c_i}\big),
  \qquad
  p_i = \softmax(Z_i),
  \label{eq:tsallis-gcn}
\end{equation}
where $c_i$ is the (possibly corrupted) training label of node $i$. What is new
relative to the i.i.d.\ case is the failure mode rather than the loss: an
unbounded $q=1$ objective does not merely memorize a mislabeled node, it
propagates that node's error to its neighbors at every aggregation step, so
label noise diffuses along edges. Because $-\lnq$ saturates on confidently
mislabeled nodes for $q<1$, the corruption stays local. Here $q$ is learned
rather than tuned, constrained to $(0.3,1.3)$ by the same smooth
reparameterization and optimized jointly with the two weight matrices on the
\emph{noisy} training objective.

\begin{lstlisting}[style=qjpy, numbers=none, float=t, caption={Node
classification with a learnable-$q$ Tsallis cross-entropy over the labeled
nodes (Eq.~\eqref{eq:tsallis-gcn}).}, label={lst:qjax-gnn}]
import jax, jax.numpy as jnp
import qjax
from qjax.nn import bounded_q

def gcn_logits(params, x, a_hat):               # a_hat = D^-1/2 (A+I) D^-1/2
    h = jax.nn.relu(a_hat @ (x @ params["w1"]))
    return a_hat @ (h @ params["w2"])

def loss_fn(params, x, a_hat, y_onehot, mask):  # mask selects labeled nodes
    q = bounded_q(params["q_raw"], 0.3, 1.3)    # q learned with the weights
    p = jnp.clip(jax.nn.softmax(gcn_logits(params, x, a_hat), -1), 1e-7, 1.0)
    ce = qjax.tsallis_cross_entropy(p, y_onehot, q=q, axis=-1)
    return jnp.sum(ce * mask) / jnp.sum(mask)
\end{lstlisting}

\begin{figure}[t]
\centering
\includegraphics[width=\textwidth]{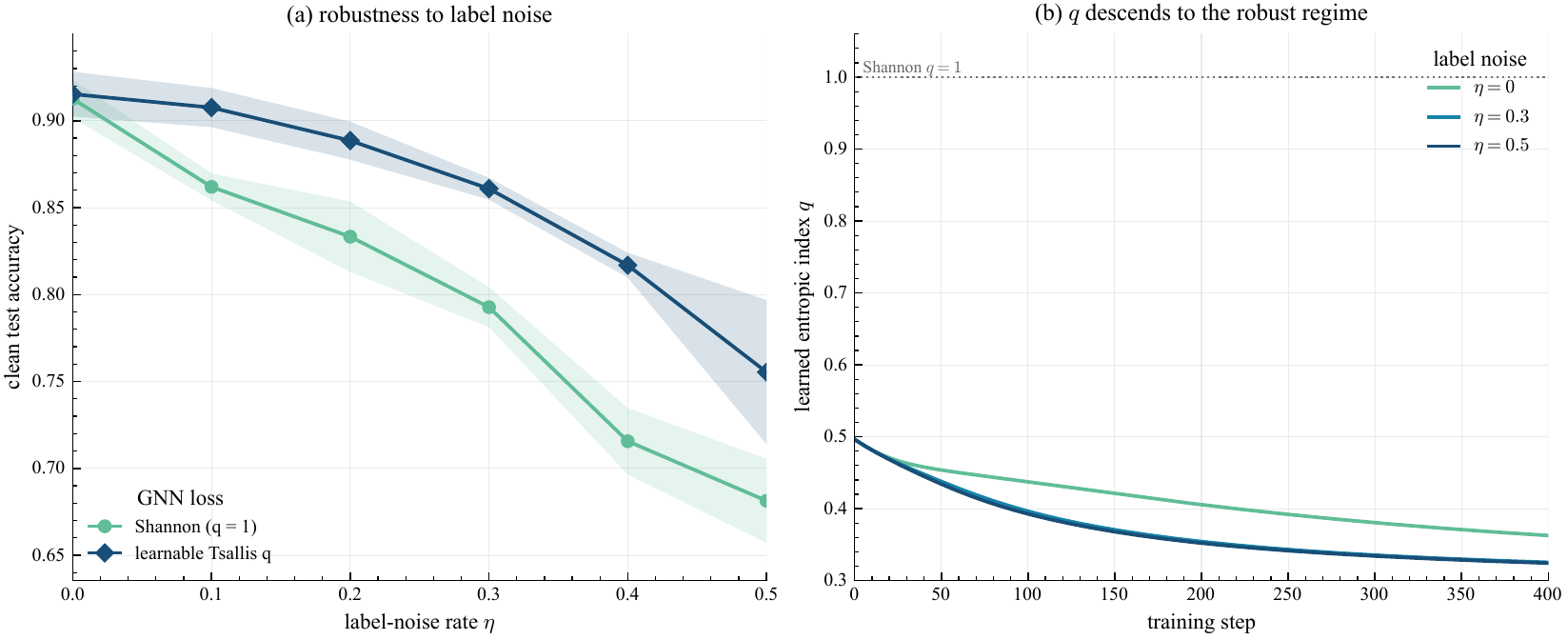}
\caption{Node classification on a stochastic block model whose communities are
the classes, under label noise (Eq.~\eqref{eq:tsallis-gcn}); bands show
$\pm1$~SEM over $4$ seeds, evaluated on the clean held-out nodes.
\textbf{(a)} Clean test accuracy versus the label-noise rate $\eta$: the two
losses are indistinguishable on clean labels ($0.91$ each), and the gap opens
steadily with contamination, reaching $8$--$10$ points from $\eta=0.2$ onward.
\textbf{(b)} The learned index during training for $\eta=0,\,0.3,\,0.5$: from an
initialization at $q\approx0.5$ it descends well below the Shannon value
(dotted line at $q=1$), and descends faster and further the noisier the labels
($q\approx0.32$ at $\eta\geq0.3$ against $q\approx0.38$ on clean labels).}
\label{fig:app-nodeclf-metrics}
\end{figure}

\begin{figure}[t]
\centering
\includegraphics[width=\textwidth]{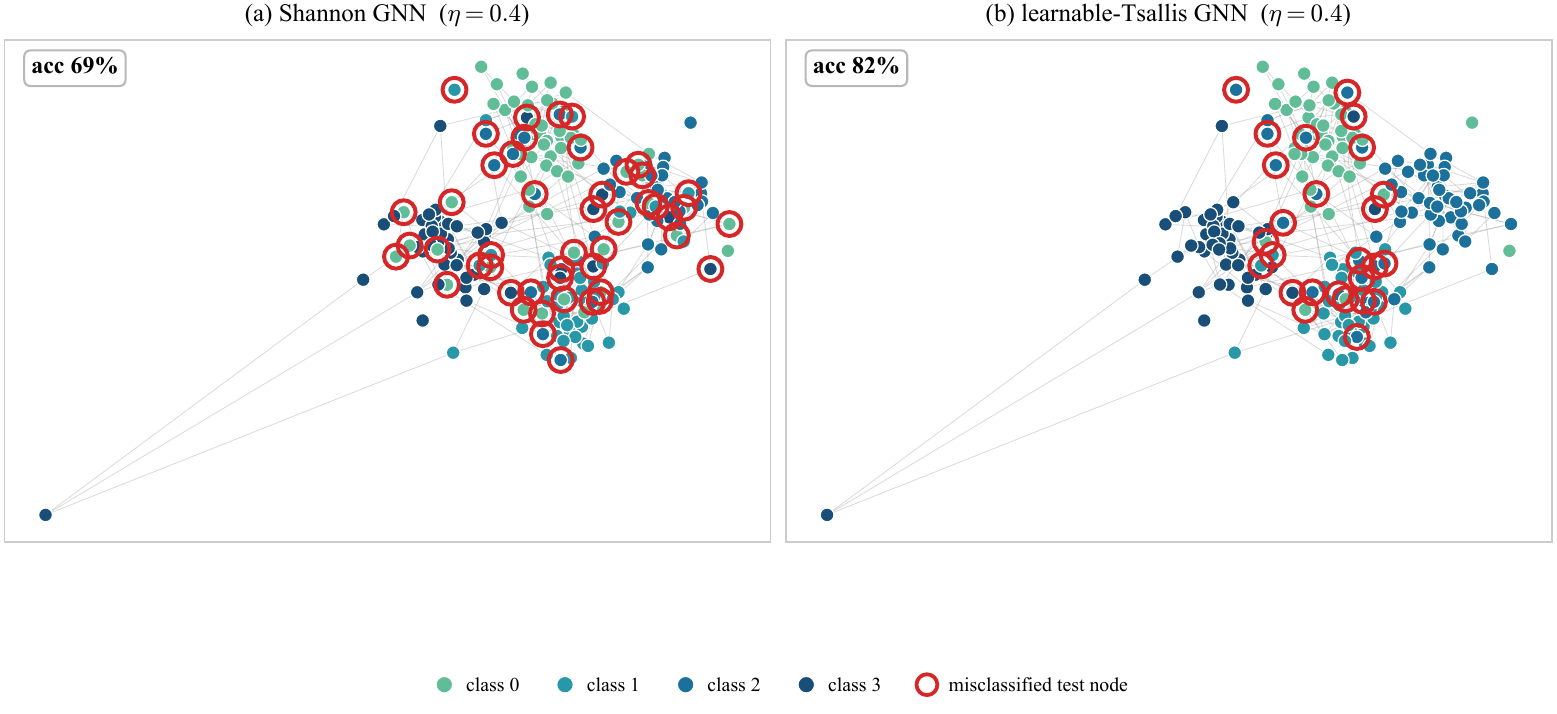}
\caption{The same graph at $\eta=0.4$, laid out by a force-directed embedding
and colored by each network's predicted class (a legible $200$-node induced
subgraph; red rings mark misclassified held-out nodes).
\textbf{(a)} The Shannon GNN ($q=1$) misreads whole stretches of two
communities, the signature of noise propagated across edges.
\textbf{(b)} The learnable-Tsallis GNN recovers the community structure, and its
residual errors sit on the boundaries between communities where the graph itself
is ambiguous.}
\label{fig:app-nodeclf-graphs}
\end{figure}

\subsubsection{Reinforcement learning}
\label{app:ex-rl}

On a $K$-armed bandit, the policy over arms is a Tsallis-entmax of learned
preferences $\mathbf{h}$, updated by the REINFORCE / gradient-bandit rule with a
running-average baseline $b$,
\begin{equation}
  \pi(a) = \operatorname{entmax}_q(\mathbf{h})_a,
  \qquad
  \mathbf{h} \;\leftarrow\; \mathbf{h} + \eta\,(r-b)\,\nabla_{\mathbf{h}}\ln\pi(a).
  \label{eq:entmax-policy}
\end{equation}
At $q=1$ this is the $\softmax$ (Boltzmann) policy, which keeps paying to sample
inferior arms; for $q>1$ clearly inferior arms are assigned exactly zero
probability. When $q$ is learnable it ascends the same return via
$\nabla_{q}\ln\pi(a)$, starting near full exploration and raising $q$ as the best
arm emerges, annealing exploration into exploitation.

\begin{lstlisting}[style=qjpy, numbers=none, float=t, caption={Tsallis-entmax bandit
policy trained by REINFORCE (Eq.~\eqref{eq:entmax-policy}).}, label={lst:qjax-rl}]
import jax
import qjax

probs = qjax.tsallis_entmax(prefs, q=q, num_iters=30)   # q=1 softmax; q>1 sparse
arm   = jax.random.choice(key, K_ARMS, p=probs)

# preferences and (optionally) q share the same REINFORCE gradient
g_prefs, g_q = jax.grad(log_prob, argnums=(0, 1))(prefs, q_raw)
\end{lstlisting}

\begin{figure}[t]
\centering
\includegraphics[width=\textwidth]{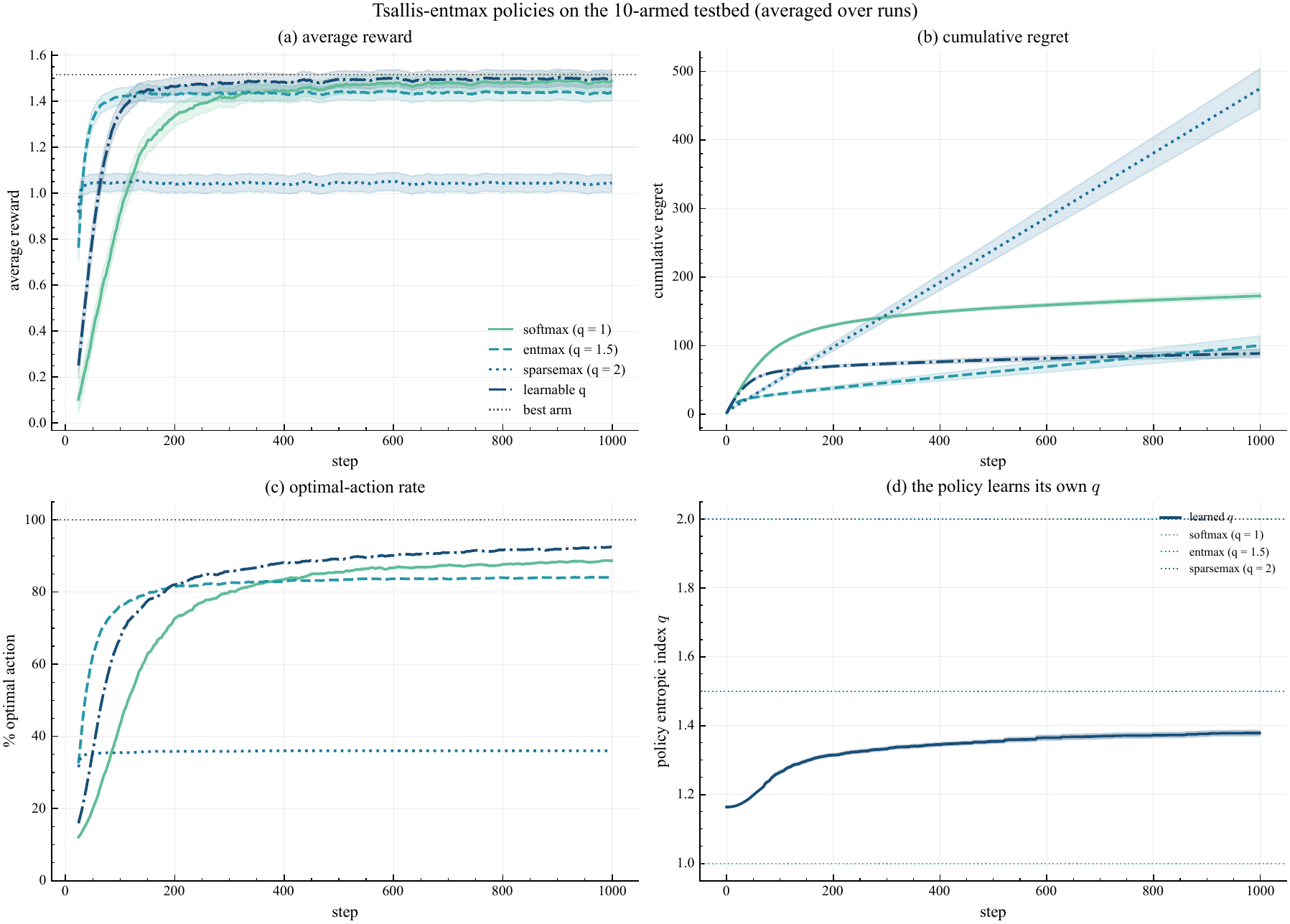}
\caption{Tsallis-entmax policies on a $K$-armed bandit
(Eq.~\eqref{eq:entmax-policy}); bands show $\pm1$~SEM. \textbf{(a)} Average
reward per step (the dashed line marks the best arm's mean value).
\textbf{(b)} Cumulative regret: the learnable-$q$ policy attains the lowest final
regret despite starting from broad exploration. \textbf{(c)} Optimal-action rate,
where $\softmax$ ($q=1$) and learnable $q$ converge to the best arm while a fixed
sparsemax ($q=2$) can lock onto a suboptimal arm prematurely. \textbf{(d)} The
index $q$ learned by the policy, rising from near-$\softmax$ toward the sparse
regime (dashed references at $q=1,\,1.5,\,2$), annealing exploration into
exploitation.}
\label{fig:app-rl}
\end{figure}

\subsubsection{Derivative-free optimization}
\label{app:ex-optimization}

A cross-entropy-method optimizer samples a population $x_i\sim\mathcal{N}(\mu,\sigma^2)$,
scores it with a cost $f$, and refits to $q$-exponentially weighted samples:
\begin{equation}
  w_i = \frac{\expq\!\big(-[f(x_i)-f_\star]/T\big)}
             {\sum_j \expq\!\big(-[f(x_j)-f_\star]/T\big)},
  \qquad
  \mu \leftarrow \sum_i w_i\,x_i,
  \quad
  \sigma^2 \leftarrow \sum_i w_i\,(x_i-\mu)^2 ,
  \label{eq:q-cem}
\end{equation}
with $f_\star=\min_i f(x_i)$ and temperature $T$. The Boltzmann weight ($q=1$)
has light, exponential tails and collapses greedily onto the current best
samples; heavy power-law tails ($q>1$) retain weight on far-from-best candidates,
sustaining the exploration needed to escape a deceptive basin.

\begin{lstlisting}[style=qjpy, numbers=none, float=t, caption={$q$-exponential
cross-entropy-method update (Eq.~\eqref{eq:q-cem}).}, label={lst:qjax-cem}]
import jax, jax.numpy as jnp
import qjax

samples = mu + sigma * jax.random.normal(key, (POP, 2))
weights = qjax.q_exp(-(costs - costs.min()) / TEMPERATURE, q)   # q=1 greedy; q>1 heavy-tailed
weights = weights / jnp.sum(weights)
mu      = jnp.sum(weights[:, None] * samples, axis=0)
sigma   = jnp.sqrt(jnp.sum(weights[:, None] * (samples - mu) ** 2, axis=0)) + 1e-3
\end{lstlisting}

\begin{figure}[t]
\centering
\includegraphics[width=\textwidth]{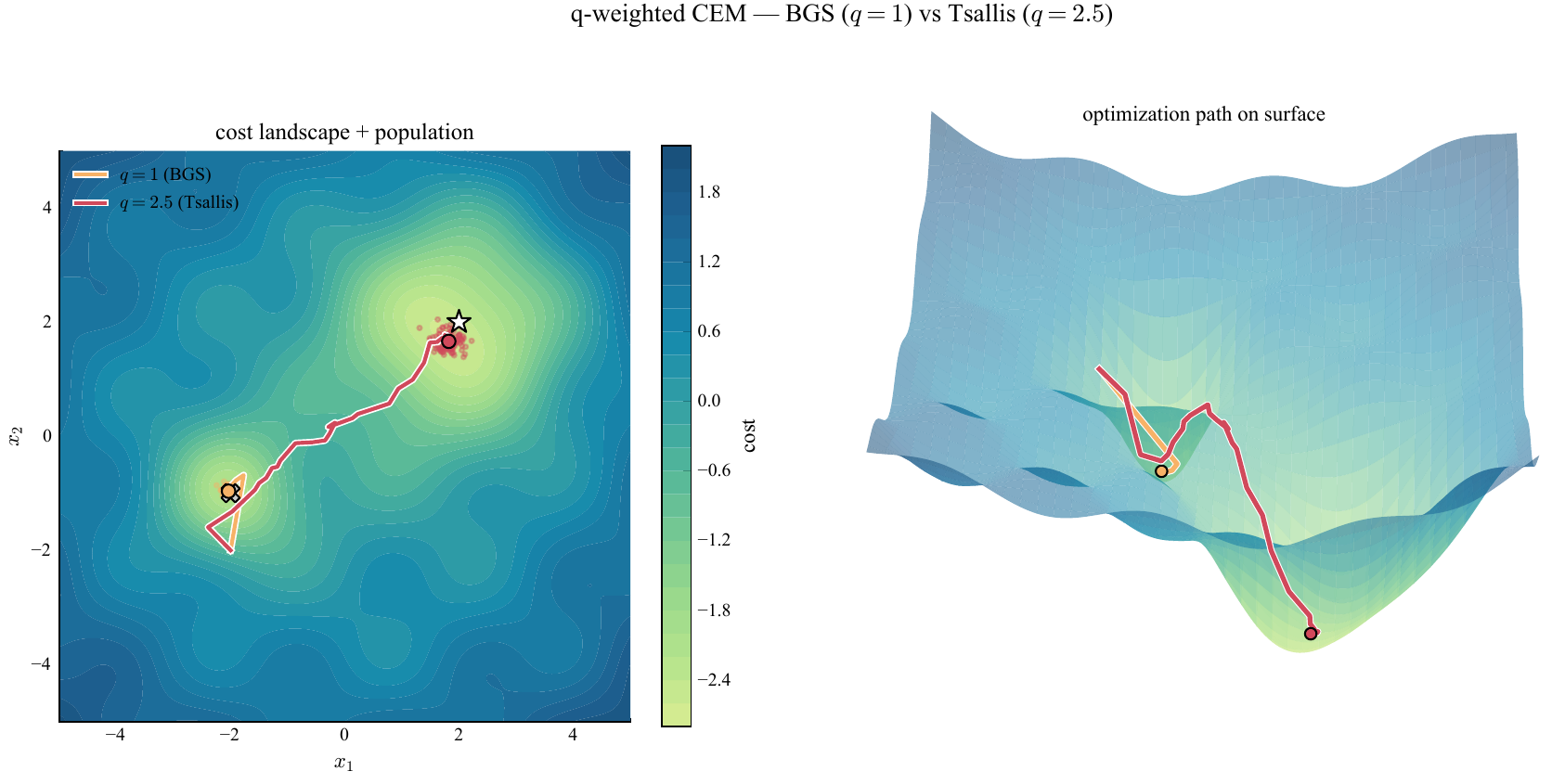}
\caption{$q$-exponential cross-entropy-method optimization on a deceptive
two-dimensional landscape (Eq.~\eqref{eq:q-cem}). \textbf{(Left)} Cost contours
with the global minimum (white star) and a shallower decoy basin ($\times$),
overlaid with the sampled population and the mean trajectories.
\textbf{(Right)} The same objective as a three-dimensional surface. The
Boltzmann weighting ($q=1$) collapses greedily into the decoy, whereas the
heavy-tailed weighting ($q=2.5$) retains mass on far-from-best candidates and
escapes to the global optimum.}
\label{fig:app-optimization}
\end{figure}

\subsubsection{Learning the entropic index}
\label{app:ex-learnq}

The central claim, that $q$ is itself differentiable, is shown directly.
Data are drawn from a $q$-Gaussian with a hidden index $q^\star$, and
$(q,\beta)$ are recovered by maximum likelihood,
\begin{equation}
  (\hat q,\hat\beta)
  = \argmin_{q,\,\beta}\ -\frac1N\sum_{n=1}^{N}\ln \mathcal{G}_q(x_n;\beta),
  \qquad q = 1 + 2\,\sigma(\theta)\in(1,3),\ \ \beta>0 ,
  \label{eq:learn-q}
\end{equation}
with the constraints enforced by a scaled sigmoid on $q$ and a softplus on
$\beta$. Plain gradient descent recovers the generative index, since
$\partial\mathcal{L}/\partial q$ flows through automatic differentiation like any
other parameter gradient.

\begin{lstlisting}[style=qjpy, float=t, caption={Learning the entropic index $q$ by
maximum likelihood: $q$ flows through \texttt{value\_and\_grad} like any other
parameter and recovers the generative $q\approx1.6$.}, label={lst:qjax-learnq}]
import jax, jax.numpy as jnp
import qjax

def neg_log_likelihood(params, x):
    beta = jax.nn.softplus(params["beta_raw"]) + 1e-3
    q    = 1.0 + 2.0 * jax.nn.sigmoid(params["q_raw"])    # constrain q to (1, 3)
    return -jnp.mean(qjax.q_gaussian_logpdf(x, q, beta))

data   = qjax.sample(jax.random.PRNGKey(0), q=1.6, beta=0.8, shape=(20_000,))
params = {"q_raw": jnp.array(0.0), "beta_raw": jnp.array(0.0)}
loss_and_grad = jax.jit(jax.value_and_grad(neg_log_likelihood))

for step in range(400):                                  # recovers q ~ 1.6
    loss, grads = loss_and_grad(params, data)
    params = {k: v - 0.05 * grads[k] for k, v in params.items()}
\end{lstlisting}

\begin{figure}[t]
\centering
\includegraphics[width=\textwidth]{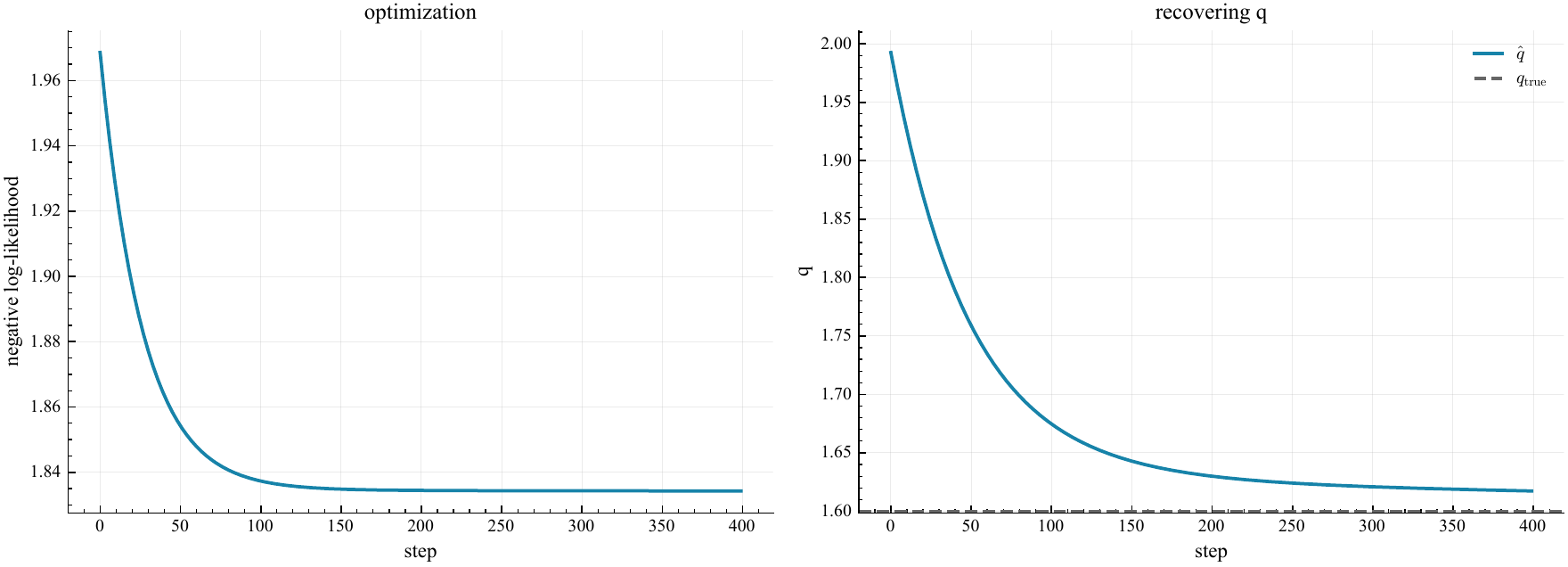}
\caption{Learning the entropic index by maximum likelihood
(Eq.~\eqref{eq:learn-q}). \textbf{(Left)} The negative $q$-Gaussian
log-likelihood decreases over training. \textbf{(Right)} The estimate $\hat q$
converges to the hidden generative index (dashed line at $q^\star=1.6$),
confirming that $q$ flows through automatic differentiation like any other
parameter.}
\label{fig:app-learnq}
\end{figure}

\end{document}